\documentclass{article}

\usepackage{arxiv}
\usepackage{newtxtext}
\usepackage{newtxmath}

\usepackage{booktabs}
\usepackage{graphicx}
\usepackage{subcaption}
\usepackage{amsmath}
\usepackage{float}
\usepackage{flafter}
\usepackage{placeins}
\usepackage{adjustbox}
\newsavebox{\crossrouteboxb}
\usepackage[backend=bibtex,style=numeric,sorting=none]{biblatex}
\usepackage[hidelinks]{hyperref}

\renewcommand{\headeright}{}
\renewcommand{\undertitle}{}
\makeatletter
\renewcommand{\@maketitle}{%
  \vbox{%
    \hsize\textwidth
    \linewidth\hsize
    \vskip 0.1in
    \@toptitlebar
    \centering
    {\LARGE\bfseries\@title\par}
    \@bottomtitlebar
    \vskip 0.1in
    \def\And{%
      \end{tabular}\hfil\linebreak[0]\hfil%
      \begin{tabular}[t]{c}\bf\rule{\z@}{24\p@}\ignorespaces%
    }
    \def\AND{%
      \end{tabular}\hfil\linebreak[4]\hfil%
      \begin{tabular}[t]{c}\bf\rule{\z@}{24\p@}\ignorespaces%
    }
    \begin{tabular}[t]{c}\bf\rule{\z@}{24\p@}\@author\end{tabular}%
    \vskip 0.4in \@minus 0.1in \center{\@date} \vskip 0.2in
  }
}
\renewcommand{\section}{%
  \@startsection{section}{1}{\z@}%
                {-2.0ex \@plus -0.5ex \@minus -0.2ex}%
                { 1.5ex \@plus  0.3ex \@minus  0.2ex}%
                {\large\bfseries\raggedright}%
}
\renewcommand{\subsection}{%
  \@startsection{subsection}{2}{\z@}%
                {-1.8ex \@plus -0.5ex \@minus -0.2ex}%
                { 0.8ex \@plus  0.2ex}%
                {\normalsize\bfseries\raggedright}%
}
\renewcommand{\subsubsection}{%
  \@startsection{subsubsection}{3}{\z@}%
                {-1.5ex \@plus -0.5ex \@minus -0.2ex}%
                { 0.5ex \@plus  0.2ex}%
                {\normalsize\bfseries\raggedright}%
}
\renewcommand{\paragraph}{%
  \@startsection{paragraph}{4}{\z@}%
                {1.5ex \@plus 0.5ex \@minus 0.2ex}%
                {-1em}%
                {\normalsize\bfseries}%
}
\makeatother

\title{Projecting Beauty: Transferring Image Aesthetic Analysis onto 3D Heritage and Urban Scenes}
\author{%
  Sophia Kovalenko \And
  Alexander M.~Rusnak \And
  Yasmine Kroknes-Gomez \AND
  Vittoria Meroni \And
  Fr\'ed\'eric Kaplan
}
\date{\'Ecole Polytechnique F\'ed\'erale de Lausanne (EPFL)}

\begin{document}
\maketitle

\begin{abstract}
Judgments about the beauty of buildings and streets inform decisions in architectural design, urban planning, and heritage conservation; yet these judgments are increasingly made by people working from scanned three-dimensional models rather than from the material fabric itself. Automatic aesthetic assessment, however, has remained almost entirely a matter of two-dimensional photographs. This paper asks what survives when the apparatus of image aesthetic assessment is carried across that dimensional gap and made to produce dense, per-point aesthetic descriptions of large heritage and urban point clouds. We examine three routes that attempt to sidestep costly, and at times ethically dubious, human-labeling campaigns at scale. The first lifts scalar scores from established image aesthetic models (NIMA, SAAN) directly onto geometry. The second trains regression heads to predict scores from the Aesthetic Visual Analysis (AVA) dataset, either on the features of frozen vision backbones (DINOv3, TIPS, SigLIP2, PE, RADSeg) or, in a 3D-native variant, on a generalist point encoder applied to feed-forward reconstructions of AVA photographs. The third queries the lifted vision--language features with a natural-language aesthetic label and reads the similarity as a score. The two supervised routes transfer poorly: heads that fit AVA well in its own domain produce class profiles on architectural scenes that neither reproduce one another nor track any other signal, and the 3D-native probe is held near chance by the geometric quality of its training reconstructions. Text querying is markedly more coherent. Once the class-text direction is projected out of each point feature, mean cross-model agreement over architectural part profiles rises from $+0.035$ to $+0.352$, and from $+0.113$ to $+0.293$ at the level of individual points, with the four backbones that pass a semantic sanity check agreeing at correlations between $0.31$ and $0.73$, while the one that fails it remains an outlier. Despite this, agreement among models is not agreement with human aesthetic perception, so we detail a part-aware rating protocol that would be useful to clarify where human and automatic rating disagree. In this work, we offer a reproducible transfer protocol, a clear account of which aesthetic pseudo-labeling routes carry signal and which do not, and a set of dense aesthetic pseudo-labels that should be read as hypotheses rather than ground-truth measurements.
\end{abstract}


\section{Introduction}
\label{sec:intro}

Three decades of digitisation have left the field of cultural heritage with an archive it cannot yet fully read. Photogrammetric surveys, laser scans, and city-scale reconstruction campaigns have produced point clouds and meshes of churches, palaces, streets, and entire historic centres, now used by conservators, planners, archaeologists, and architectural historians. These 3D models can reveal which facades will reward restoration, which street frontages should be protected, and which views into a square are worth defending against new construction. Each question involves an aesthetic judgment at the scale of a moulding or cornice rather than a whole building. At best, computational tools encode material classes, geometric primitives, and semantic labels, while very little in the standard pipeline produces anything a humanities scholar would recognise as an aesthetic description.

In lieu of human ratings or strong multimodal 3D encoders, the obvious move is to take a deep learning model that scores photographs for beauty and point it at a rendered view of a scanned building. Image aesthetic assessment was built on photographic critique communities, whose criteria concern exposure, framing, subject isolation, and the conventions of the well-made picture. Architectural aesthetics has a longer and different history. From Vitruvius onward it has treated beauty as a property distributed across the parts of a construction and bound up with their fitness, proportion, and material honesty \cite{vitruvius1914tenbooks,alberti1988art,ruskin1849sevenlamps}, while heritage theory has insisted that a monument's values are plural and need not coincide with contemporary taste \cite{riegl1982monuments,brandi2005theory,nara1994authenticity}. Transferring photographic taste onto architectural geometry is therefore a translation between two aesthetic cultures. This differs from perceptual quality assessment of 3D data, which asks whether a model faithfully records its source rather than whether its surfaces are beautiful \cite{10415439,8803298,laazoufi2025pointcloudqualityassessment}.

We approach this through dense, per-point pseudo-labels rather than a single number per scene, since a scene-level score cannot delineate the aesthetic effect of the ornament above a portal in contrast to the pavement below it. This distributed, particular form of statement has long been central to architectural and heritage scholarship. Without generalist language--3D multimodal models, producing per-point scores requires a bridge from images into geometry. We build this bridge with feature lifting: virtual cameras are placed around a scene, a frozen 2D model is run on each rendered view, and the resulting pixel-aligned signals are projected back onto the points that view can see \cite{peng2023openscene3dsceneunderstanding,opencity3d2025,wang2025openurban3dannotationfreeopenvocabularysemantic}. What is lifted can be a scalar aesthetic score or an open-vocabulary semantic vector, and that choice defines the routes to aesthetic assessment we compare.

\begin{figure}[tbp]
    \centering
  \includegraphics[width=\textwidth]{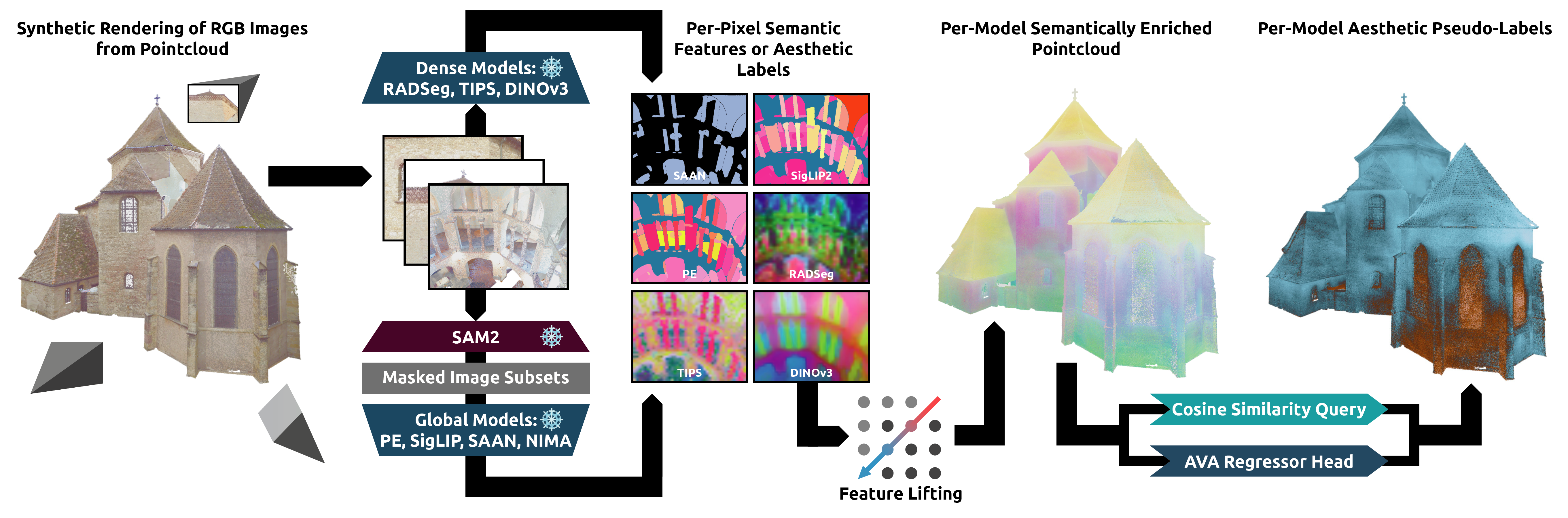}
  \caption{Rendering and feature lifting from frozen 2D models to per-point aesthetic pseudo-labels. Global models pass through mask-then-embed; dense models contribute patch features directly.}
  \label{fig:pipeline}
\end{figure}

The first route lifts scalar scores: NIMA and SAAN return one aesthetic score per processed image, so each base image is first divided into regions using Segment Anything Model 2 (SAM 2) before each region is scored separately to produce dense aesthetic scores \cite{NIMA,yi2023artisticimageaestheticsassessment,ravi2024sam2}. The second learns a mapping by training identical regression heads on AVA \cite{6247954} over five frozen vision backbones, with a 3D-native counterpart probing a generalist unimodal point encoder \cite{zhang2026utoniaencoderpointclouds} over 3D-reconstructed AVA photographs \cite{keetha2026mapanythinguniversalfeedforwardmetric}. The third uses natural-language querying, comparing a phrase such as ``beautiful object'' against each lifted point feature without training.

The routes do not fare equally. AVA regression transfers poorly, lifted IAA scalars agree partly because they share a segmentation stage, and text querying produces the most coherent signal after class residualisation.

Section~\ref{sec:related} situates the work in image aesthetics, architectural heritage, and feature lifting. Sections~\ref{sec:method} and~\ref{sec:results} describe the pipeline and its results, while Section~\ref{sec:discussion} states what those results license.

\section{Related Work}
\label{sec:related}

\subsection{Image aesthetic assessment}
\label{sec:rel-iaa}

Computational aesthetics of photographs began with hand-designed measurements of properties such as exposure balance, color distribution, subject sharpness, and compositional structure \cite{IAA_models_comparative_study}. These features were interpretable and brittle in equal measure, and were largely displaced once convolutional networks could learn the relevant regularities directly from rated corpora \cite{IAA_models_comparative_study,Wang,Valenzise_Kang_Dufaux_2021}. NIMA marked a decisive step, predicting the full distribution of human ratings for an image rather than a single mean, which let it express how contested a judgment was as well as how favorable \cite{NIMA}. Yi et al.\ addressed artworks with the Style-specific Art Assessment Network, pairing general and style-conditioned pathways and introducing the Boldbrush Artistic Image Dataset (BAID) on which it was trained \cite{yi2023artisticimageaestheticsassessment}.

Progress here has been driven by datasets at least as much as by architectures. AVA remains the reference corpus, with more than 250{,}000 photographs each carrying a distribution of scores from members of a critique community \cite{6247954}, joined by BAID, the aesthetic subsets of LAION-5B, the Reddit Photo Critique Dataset, and the Aesthetics of Paintings and Drawings Dataset \cite{yi2023artisticimageaestheticsassessment,LAION-Aesthetics,nieto2022understandingaestheticslanguagephoto,jin2024apddv2aestheticspaintingsdrawings}. Because human ratings are expensive, pseudo-labelling with large models has recently become an accepted way of extending them in 2D and 3D \cite{10746353,xu2026aes3daestheticassessment3d}.

The prior finding most relevant to the present study is that generalist models never trained on explicit aesthetic labels are often better aesthetic predictors than models that were. A substantial line of research reports that features from CLIP-family vision--language encoders support aesthetic prediction matching or exceeding purpose-built IAA systems across photographic, artistic, and mixed benchmarks \cite{10.3389/frai.2022.976235,aesclip1,aesclip2,aesclip3,aesclip4,aesclip5,aesclip6,aesclip7}, and the effect is not confined to models with a language channel: self-supervised encoders in the DINO family, trained with no captions and no human annotation of any kind, likewise yield representations from which aesthetic judgments can be decoded with high fidelity \cite{aesdino1,aesdino2,aesdino3,aesdino4,dino}. Conwell et al.\ put the question directly, training linear decoders to predict human beauty ratings from unimodal vision embeddings, unimodal language embeddings, and multimodal combinations; the visual modality accounts for most of the explained variance, while language and multimodal alignment add comparatively little \cite{conwell2024usingmultimodaldeepneural}.

The convergent implication bears directly on this study. Encoders trained independently, on different data and objectives, and in some cases without language at all, can recover the same human ratings of images with high fidelity. A considerable portion of aesthetic variance is therefore carried by visual structure that many learning procedures capture whether or not they were pointed at aesthetics. This claim concerns statistical recoverability, not the metaphysics of beauty, but it justifies using generalist backbones as aesthetic instruments and including DINOv3 alongside the vision--language models we test in this study \cite{dinov3}.

\subsection{The aesthetics of architecture and heritage}
\label{sec:rel-heritage}

Architectural theory has treated beauty as a technical subject for as long as it has been written down, and has consistently located that beauty in the relation between parts rather than in the impression made by a whole. Vitruvius sets \emph{venustas} beside \emph{firmitas} and \emph{utilitas} as one of three criteria a building must satisfy, placing delight on the same footing as structural soundness and fitness for purpose rather than treating it as an ornamental afterthought \cite{vitruvius1914tenbooks}. Alberti develops this into an account in which beauty consists in a reasoned harmony of the parts, such that nothing could be added, removed, or altered except for the worse, and in which ornament is a subordinate addition governed by decorum: what suits a temple does not suit a warehouse \cite{alberti1988art}. Ruskin, writing against industrial production, ties architectural value to the visible labour and material honesty of surfaces, reading mouldings, carving, and weathering as bearers of moral and historical meaning rather than as decoration applied over a structure \cite{ruskin1849sevenlamps}.

Three consequences of this tradition matter for computational work. The first is that architectural aesthetic judgment is intrinsically part-aware: when Alberti discusses concinnitas or Ruskin the carved capital, the unit of judgment is a member of a building rather than a view of it, which is exactly the granularity a dense per-point representation can express and a scene-level score cannot. Second, architectural judgment depends on appropriateness: decorum relates an element to its building type and setting. A scoring function that assigns beauty to surfaces without considering what they belong to therefore answers a different question from the one posed by this tradition. The third is that these judgments are bound to material presence. Pallasmaa's critique of ocularcentrism holds that architectural experience is haptic and kinaesthetic before it is pictorial, and that architecture designed for the photograph loses the qualities that reward inhabitation \cite{pallasmaa2005eyes} while Scruton argues that architecture is an aesthetic of everyday life in which fit with a street matters more than isolated visual incident \cite{scruton2013aesthetics} and Carlson makes the corresponding point for landscape \cite{carlson1979appreciation}. Tzonis and Lefaivre likewise argue that buildings should be understood within the architectural systems of an urban landscape. They introduce the concept of \textit{parataxis}, which evaluates individual building elements in semi-exocentric succession, as they would be experienced while walking through a street, rather than by their harmony from a single vantage point or within an isolated facade \cite{poeticsoforder}. A system which evaluates the summation of individual rendered views of a building works obliquely to or at a distance from these hermeneutics of evaluation.

Heritage theory further complicates this picture: beauty is only one of a monument's values. Riegl distinguishes age, historical, commemorative, and newness values, which may conflict when weathering is valued by one observer and treated as deterioration by another. Brandi, Nara, and Smith likewise show that heritage judgments involve competing historical, aesthetic, cultural, and institutional perspectives \cite{riegl1982monuments,brandi2005theory,nara1994authenticity,smith2006uses}.

These works explain why we compute scores per point and per class on our 3D scenes, why photographic ratings cannot be treated naively as heritage value, and why disagreement between computational regimes is not automatically a defect. The considerable subtlety and complexity of architectural and heritage value judgments also motivate the move from semantically limited closed-set understanding to open-vocabulary representations, whose vectors can encode concrete qualities such as materiality and age as well as abstract concepts such as beauty and historical significance under the right conditions.

\subsection{Aesthetic assessment beyond the picture plane}
\label{sec:rel-3d}

Compared with IAA, work on 3D aesthetics is comparatively sparse and often relies on questionnaires that do not scale, proxy labels that are only loosely aesthetic, or object-centric datasets. Zhu et al.\ score residential landscape areas from 3D city models by questionnaire \cite{su141811500}; Angelini et al.\ use SketchFab likes and views for ViDA 3D, confounding quality with popularity in a collection dominated by objects \cite{Angelini2020ViDA3D}; and Gulhan et al.\ show that judgments of 3D artworks shift between virtual reality and online presentation \cite{Gulhan2022Aesthetic3D}. Existing work also predicts preferences for abstract 3D shapes, while AES3D scores Gaussian Splatting scenes from rendered views without lifting features onto points \cite{devlau2017polygons,devlau2022learning,devlau2023comparing,xu2026aes3daestheticassessment3d}. No public dataset offers dense human aesthetic annotation of heritage or urban geometry, which makes pseudo-labelling necessary but also limits the strength with which those pseudo-labels can claim to assess a scene's aesthetic content.

\subsection{Foundation models and feature lifting}
\label{sec:rel-lifting}

The foundation models we use as instruments come from two families: global and dense encoders. CLIP established joint image-text training, allowing any phrase to act as a classifier without further training \cite{clip}. SigLIP2 retains this design with a sigmoid objective, scaled data, and multilingual coverage \cite{zhai2023sigmoidlosslanguageimage,tschannen2025siglip2multilingualvisionlanguage}, while the Perception Encoder provides a contemporary alternative for visual embeddings \cite{bolya2025perceptionencoderbestvisual}. The DINO family instead learns by self-distillation across augmented views, producing high-resolution feature maps without text or labels \cite{dino,dinov2,dinov3}. Language alignment remains an active problem: \emph{dino.txt} aligns DINO features to text using whole-image captions, while Talk2DINO grounds them at pixel and mask level \cite{dinotxt,talk2dino}. RADIO, the backbone of RADSeg, distils several teachers to combine dense and language-aligned strengths, while TIPS optimises patch-level language binding \cite{ranzinger2024amradioagglomerativevisionfoundation,alama2026radsegunleashingparametercompute,cao2026tipsv2advancingvisionlanguagepretraining}.

Carrying these representations into 3D has become a relatively standard procedure. OpenScene projects per-pixel features onto points using camera geometry and depth, while mask-then-embed pipelines extend the idea to models that emit one embedding per image \cite{peng2023openscene3dsceneunderstanding,opencity3d2025,qin2024langsplat3dlanguagegaussian}, and recent systems apply the procedure at city scale, rendering the views they need where capture imagery is absent \cite{wang2025openurban3dannotationfreeopenvocabularysemantic}. On the 3D side, Sonata and Concerto improved point representation learning through joint 2D and 3D pretraining, and Utonia consolidates these advances into a cross-domain encoder \cite{sonata,zhang2026concertojoint2d3dselfsupervised,zhang2026utoniaencoderpointclouds}. Every component listed here was designed to transport general and flexible semantics. What we examine is what happens when the quantity transported or queried is an aesthetic judgment instead of a class name.

\section{Method}
\label{sec:method}

\subsection{Datasets and their usages}
\label{sec:data}

The datasets determine both how we use them and what can be validated. \textbf{ArCH}, \textbf{Turin3D}, and \textbf{YTU3D} carry semantic labels only, so their aesthetic maps are pseudo-labels, while \textbf{TerraSky3D} supports statements about rendering fidelity. \textbf{AVA} is the only source of explicit human aesthetic judgments, and those judgments concern whole photographs.

ArCH contains annotated historical-building point clouds with architectural part labels \cite{arch}; its source images and camera poses are not distributed, requiring synthetic rendering. Turin3D covers roughly $1.43\,\mathrm{km}^{2}$ of central Turin with close to 70 million aerial LiDAR points \cite{barco2025turin3devaluatingadaptationstrategies}. YTU3D is an aerial urban dataset with 45 semantic classes, used only for the semantic check \cite{ytu3d}. TerraSky3D provides landmark reconstructions with photographs, poses, and depth \cite{durso202Xterrasky3d}.

\subsection{Synthetic rendering for scenes without capture imagery}
\label{sec:render}

\begin{figure}[tbp]
\centering
  \includegraphics[width=\textwidth]{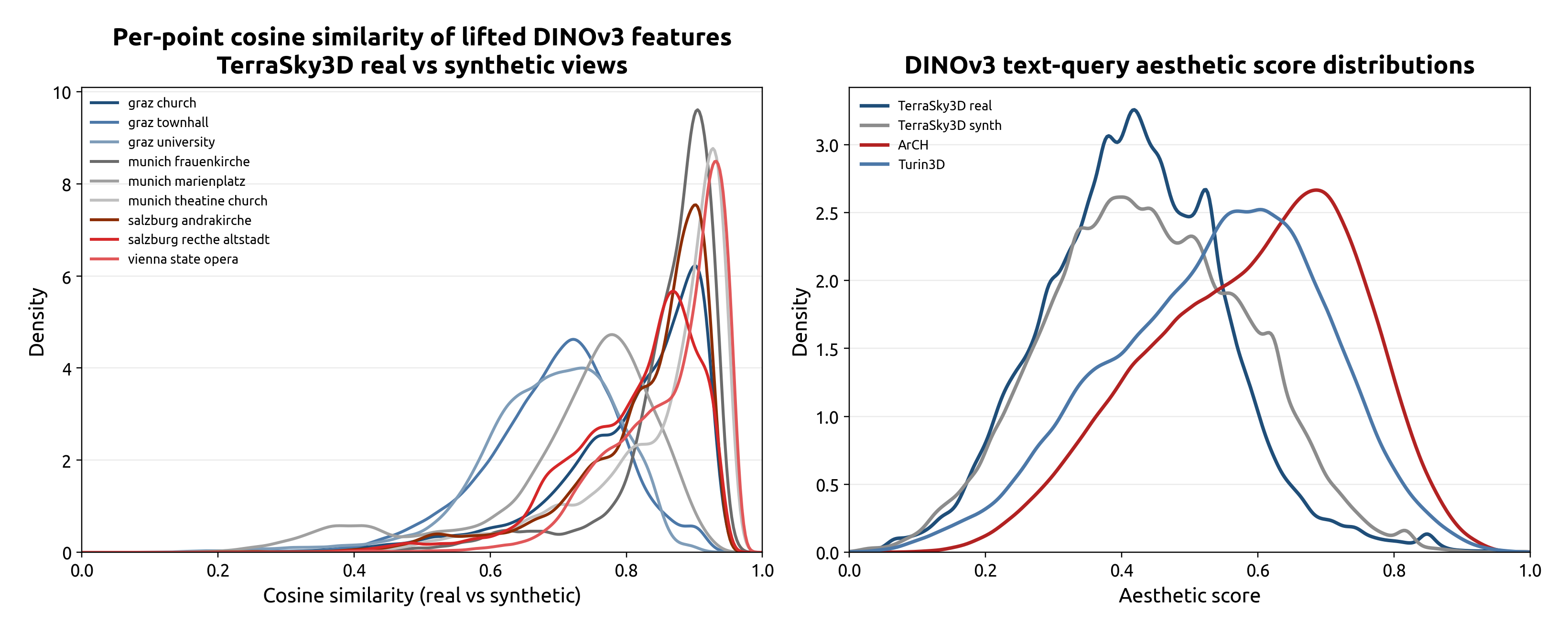}
  \caption{Real versus synthetic lifting on TerraSky3D at matched camera positions; rendered views preserve most transferable signal.}
  \label{fig:synth}
\end{figure}

Every 2D model takes an image as input, while many heritage point clouds arrive without any. We therefore place virtual cameras around each scene and render RGB views with an offscreen renderer using the color carried on the points. TerraSky3D provides real images with known poses for a direct comparison: at matched camera positions, real and synthetic lifts have high per-point cosine similarity for the stronger backbones and aesthetic score distributions separated by a Wasserstein distance of $0.01$ (Figure~\ref{fig:synth}). Renders are therefore usable where photographs are unavailable, with only a residual domain shift.

\subsection{From images to per-pixel signals}
\label{sec:mask}

Modern vision transformers divide an image into patches and produce a vector for each, encoding what the model perceives at that location in context. DINOv3, TIPS, and RADSeg expose these patch vectors directly, so a pixel-aligned feature map is available.

On the other hand, SigLIP2 and PE return one semantic vector per image, while NIMA and SAAN return one scalar aesthetic score, which would paint every visible point with the same value after projection. We recover spatial variation by using SAM 2 to propose regions, passing each region to the model, and writing the resulting embedding or score back to its pixels \cite{ravi2024sam2}. The resulting piecewise-constant map is controlled by the segmenter's granularity, a dependency that matters for Section~\ref{sec:res-iaa}.

\begin{figure}[tbp]
\centering
  \begin{subfigure}{0.327\textwidth}
    \includegraphics[width=\linewidth]{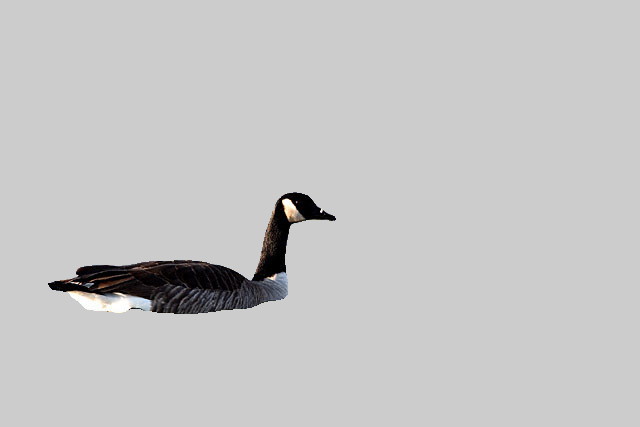}
      \caption{Strict mask}
  \end{subfigure}\hfill
  \begin{subfigure}{0.327\textwidth}
    \includegraphics[width=\linewidth]{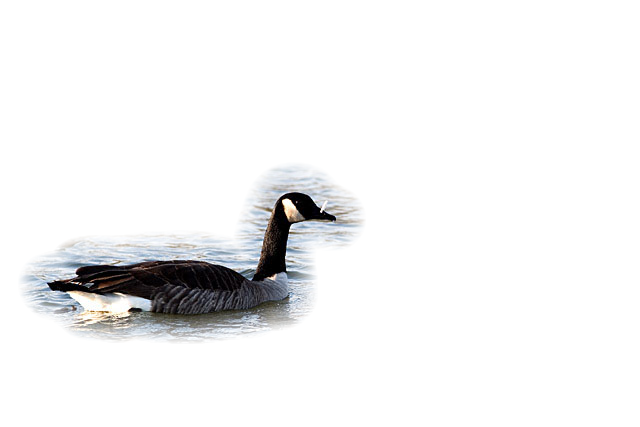}
        \caption{Faded context}
  \end{subfigure}\hfill
  \begin{subfigure}{0.327\textwidth}
    \includegraphics[width=\linewidth]{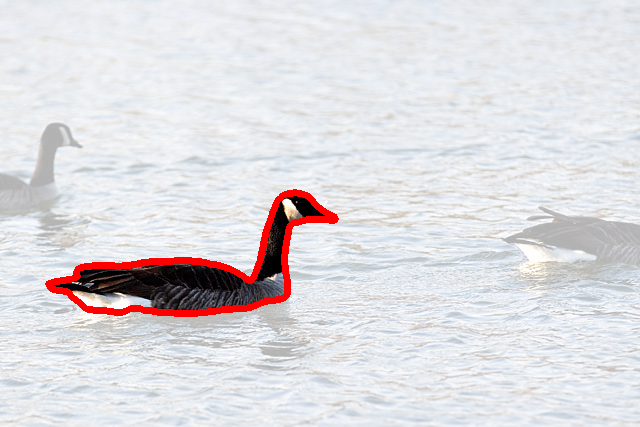}
        \caption{Red border}
  \end{subfigure}
  
  \caption{The three masking strategies compared for models that return a single value per image. The red border is retained on the evidence of Table~\ref{tab:mask-ablation}.}
  \label{fig:masks}
\end{figure}

This style of mask$\rightarrow$embed$\rightarrow$assemble$\rightarrow$lift pipeline often fades the surrounding image or adds a colored border so the encoder retains figure--ground cues \cite{opencity3d2025,qin2024langsplat3dlanguagegaussian}. To validate these masking strategies for dense IAA, we compare a strict mask, faded context, and a red border over a faded background (Figure~\ref{fig:masks}). On a common AVA subset, the red border gives the strongest NIMA alignment, Spearman $0.378$, and the strongest PE text agreement, so we adopt it for global models. SAAN correlates with AVA at essentially zero under every operator. We interpret this as a limit of transferring an artwork-oriented scorer to photographic crops, and as evidence of the fragility of specialist IAA models under domain shift.

\begin{table}[tbp]
\centering
\caption{Masking ablation against AVA on a common image subset, averaged over instance masks. Sp and P denote Spearman and Pearson correlation with the AVA mean. PE and SAAN use different native scales and are compared by rank correlation only.}
\label{tab:mask-ablation}
\begin{tabular}{lcc cc cc}
\toprule
& \multicolumn{2}{c}{PE} & \multicolumn{2}{c}{NIMA} & \multicolumn{2}{c}{SAAN/BAID} \\
\cmidrule(lr){2-3}\cmidrule(lr){4-5}\cmidrule(lr){6-7}
Ablation & Sp & P & Sp & P & Sp & P \\
\midrule
Original & 0.209 & \textbf{0.201} & 0.198 & 0.214 & 0.004 & 0.010 \\
Faded context & 0.193 & 0.191 & 0.266 & 0.281 & $-0.009$ & $-0.007$ \\
Red border & \textbf{0.211} & 0.195 & \textbf{0.378} & \textbf{0.396} & \textbf{0.035} & \textbf{0.032} \\
\bottomrule
\end{tabular}
\end{table}

\subsection{Lifting signals onto geometry}
\label{sec:lift}

With a per-pixel signal available for every rendered view, the transfer into 3D is a deterministic mapping. Each point collects candidate values from the cameras that see it. We average these values with weights based on how directly each camera faces the estimated surface normal, so frontal views count more than grazing views.

This is not a `true' route to viewpoint-invariant aesthetics. A photographic aesthetic score is a judgment about a view, conditioned on framing, foreshortening, and the lighting of the synthetic camera, and averaging many such judgments produces a visibility-weighted summary rather than a measurement of an intrinsic property of the surface. The dense maps in this paper are therefore unable to escape the accumulated biases of their viewpoints, even if views from a multitude of locations around a 3D object can in some sense approximate a view-independent perspective. The need for truly exocentric, view-independent aesthetic understanding of 3D objects or buildings motivates our tests with a 3D point-encoder backbone and the future development of stronger 3D foundation models.

\subsection{Routes to 3D aesthetic scores}
\label{sec:scoring}

\paragraph{Lifting scalar aesthetic scores.}
The most direct route lifts numbers rather than vectors. NIMA and SAAN are run on each masked region, their scores are written back to pixels, and the pixels are projected onto points exactly as features would be. NIMA is trained on AVA and encodes photographic taste; SAAN is trained on BAID and encodes an artwork-oriented one. Both are older specialist systems relative to current backbones, which is what makes them useful as controls: were mature IAA heads already capable of consistent dense aesthetics on architecture, that would appear here as coherent and mutually aligned maps which distinguish strongly between semantic objects.

\paragraph{Natural language querying.}
Because vision--language backbones place images and text in a single space, a phrase can be encoded and compared against a lifted point feature without any training. We encode the prompt ``beautiful object'', take its cosine similarity with each point's lifted feature, and map the result into $[0,1]$. Nothing is fitted to an aesthetic dataset; the score reflects only how the backbone's pretraining has arranged words and pictures relative to one another in its embedding geometry. The same lifted features can be queried with any text, including the dataset's class-label names, to measure open-vocabulary semantic segmentation; we use this as a check on the quality of the lifted features (Section~\ref{sec:res-ovss}).

\paragraph{Training a regression head with AVA.}
In an attempt to translate every backbone into one aesthetic ``vocabulary'', we train identical shallow multi-layer perceptrons on AVA over frozen features, using patch embeddings for DINOv3, TIPS, and RADSeg and mask embeddings for PE and SigLIP2, with architecture and optimisation held fixed so that differences reflect the representation rather than the head (Appendix~\ref{app:train}). Each patch or mask inherits the global score of its source photograph. The resulting head can learn visual patterns associated with highly rated photographs, but its weak localisation supervision does not specify which architectural parts deserve a score. Local variation is therefore an emergent effect of features, masking, and aggregation rather than a traditionally supervised prediction.

\paragraph{Probing a 3D encoder directly.}
Every route above remains mediated by rendered views and is therefore ultimately view-dependent. As a complement and step toward 3D-native, exocentric aesthetic assessment, we reconstruct AVA photographs into point clouds with MapAnything \cite{keetha2026mapanythinguniversalfeedforwardmetric} before training an attentive probe \cite{psomas2026attentionpleaserevisitingattentive} with PointRoPE positional encoding on frozen Utonia features. Analogously to pixel inheritance for the image regressors, this formulation assigns every point the global AVA score of its source photograph. Figure~\ref{fig:mapanything-cat} gives a visual example of the reconstruction distortion and of the locally distinct predictions that can nevertheless emerge under this global-supervision probing paradigm.

\subsection{Reading the maps with aesthetic ratios}
\label{sec:ratio}

Dense scores are hard to compare across models by eye, and scene-level scores lack specificity. We therefore summarise each map against the semantic labels the dataset already provides. For every class we compute
\begin{equation}
\mathrm{Aesthetic\ Ratio}
=
\frac{\#\{\mathrm{points\ with\ score}>0.5\}}{\#\{\mathrm{points\ in\ class}\}},
\end{equation}
giving each model, on each scene, a short profile over classes such as decoration, column, vault, and floor, which allows two models to be compared by correlation. The threshold is fixed rather than set to the mean of each scene, which would split every scene into equal halves by construction and make a uniformly plain building indistinguishable from a splendid one. The measure is not a ranking of model quality: a model that paints an entire city above $0.5$ obtains high ratios everywhere, which reflects an absence of discrimination rather than an abundance of beauty.

\section{Results}
\label{sec:results}

\subsection{Semantic structure survives the lift}
\label{sec:res-ovss}

Before evaluating aesthetics with our lifted features, it is worth establishing that the geometric transfer works at all. Querying the lifted features with dataset class names and comparing against the provided labels gives the accuracies and intersection-over-union scores in Table~\ref{tab:miou}. DINOv3 with Talk2DINO grounding is clearly the strongest, reaching $19.61$ mIoU on ArCH and $13.39$ on Turin3D, with PE and TIPS producing weaker but non-trivial structure and RADSeg failing almost completely, at below one point of mIoU on ArCH and effectively zero on Turin3D. These values are modest in absolute terms, as is usual for zero-shot open-vocabulary segmentation this far from the pretraining distribution, but they establish that recognisable semantic structure reaches the point cloud for four of the five backbones. Their usefulness is that they give an independent ordering of backbone quality, against which the aesthetic agreement of Section~\ref{sec:res-residual} can be compared. We include RADSeg in the correlation analysis for the sake of completeness, but exclude it from the text query ensembles because of its poor semantic segmentation performance. 

\begin{table}[tbp]
\centering
\caption{Open-vocabulary semantic segmentation of lifted features against dataset class labels (mAcc and mIoU, per cent). This is a check on geometric transfer, not an aesthetic evaluation.}
\label{tab:miou}
\begin{tabular}{lcc cc cc}
\toprule
& \multicolumn{2}{c}{ArCH} & \multicolumn{2}{c}{Turin3D} & \multicolumn{2}{c}{YTU3D} \\
\cmidrule(lr){2-3}\cmidrule(lr){4-5}\cmidrule(lr){6-7}
Model & mAcc & mIoU & mAcc & mIoU & mAcc & mIoU \\
\midrule
PE & 24.67 & 9.72 & 31.19 & 6.60 & -- & -- \\
SigLIP2 & 7.35 & 1.16 & 32.43 & 5.55 & -- & -- \\
RADSeg & 5.50 & 0.95 & 0.15 & 0.18 & 0.85 & 0.24 \\
TIPS & 7.49 & 2.13 & 23.05 & 6.87 & 22.48 & 3.47 \\
DINOv3 & \textbf{43.19} & \textbf{19.61} & \textbf{33.06} & \textbf{13.39} & \textbf{42.86} & \textbf{5.15} \\
\bottomrule
\end{tabular}
\end{table}

\subsection{Regression from AVA does not survive the change of domain}
\label{sec:res-regress}

Though the image regression heads effectively predict AVA scores, they do not accurately carry this performance to 3D lifted features. Table~\ref{tab:regressors} reports held-out performance on AVA against a constant-mean baseline: DINOv3 and TIPS both reduce mean squared error by more than half while correlating with human ratings at $0.7570$ and $0.7312$, SigLIP2 and PE follow at $0.6930$ and $0.6337$, and only RADSeg fails outright at $0.1956$, consistent with its collapse in Table~\ref{tab:miou}. On its own terms the table confirms the literature reviewed in Section~\ref{sec:rel-iaa}: generalist backbones, none built for aesthetics, support aesthetic prediction competently, and the two with the strongest dense features support it best.

\begin{table}[tbp]
\centering
\caption{AVA regression heads on held-out photographs, against a constant-mean baseline. The heads fit the photographic domain well; Section~\ref{sec:res-regress} shows that this does not transfer.}
\label{tab:regressors}
\begin{tabular}{lcccccc}
\toprule
Metric & Baseline & PE & SigLIP2 & RADSeg & TIPS & DINOv3 \\
\midrule
MSE $\downarrow$ & 0.5349 & 0.3224 & 0.2818 & 0.5143 & 0.2489 & \textbf{0.2295} \\
MAE $\downarrow$ & 0.5760 & 0.4418 & 0.4131 & 0.5642 & 0.3883 & \textbf{0.3744} \\
RMSE $\downarrow$ & 0.7314 & 0.5678 & 0.5309 & 0.7171 & 0.4989 & \textbf{0.4790} \\
Corr.\ $\uparrow$ & -- & 0.6337 & 0.6930 & 0.1956 & 0.7312 & \textbf{0.7570} \\
\bottomrule
\end{tabular}
\end{table}

What does not follow is that the heads remain meaningful once they leave AVA. On rendered heritage and urban views, their class profiles neither converge on one another nor reproduce the profiles produced by text querying, as the right-hand panels of Figure~\ref{fig:radar} show. A shared photographic label space therefore does not induce a shared aesthetic geography.

The reasons are not mysterious. AVA is narrower than the corpora on which these backbones were pretrained. Furthermore, a head fitted to a photographic aesthetic distribution learns what a well-exposed and well-composed photograph looks like, while a synthetic render of a laser-scanned vault often violates those expectations. Feature averaging during lifting may also move the embeddings farther from the regions associated with natural photographs.

\begin{figure}[H]
\centering
  \includegraphics[width=0.50\textwidth]{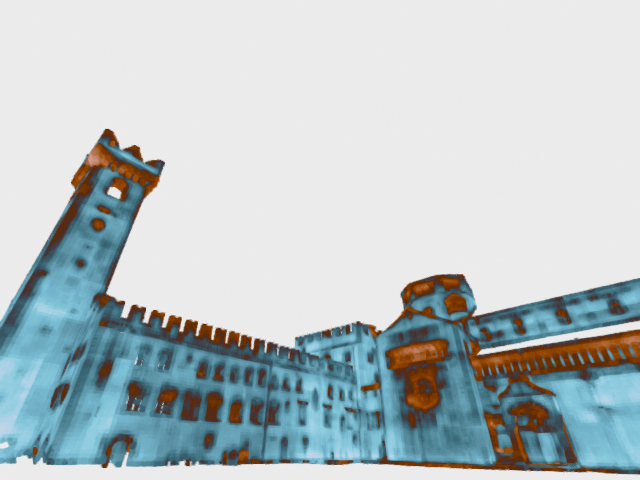}
  \caption{Results from the 3D-native aesthetic probe. Qualitative assessment of the Utonia+AVA adapter aesthetic map shows clear semantic salience. Despite this, separability is marginal on the AVA reconstruction evaluation set, which bounds what can be claimed about aesthetic signal recoverable from geometry under global labels on distorted reconstructions.}
  \label{fig:utonia}
\end{figure}

The 3D-native probe is bounded by the same logic and by unreliable geometry. Trained on Utonia features over MapAnything reconstructions of AVA photographs, it reaches a held-out ROC-AUC of $0.5594$ (Figure~\ref{fig:utonia}), above chance but only marginally separable. The probe must find aesthetic structure in a distorted representation using labels that describe the photograph rather than any part of it. We report this as a bound on the present pathway, not as evidence that native 3D aesthetics are out of reach.

\subsection{Lifted IAA scalars agree, but the segmentations drive convergence}
\label{sec:res-iaa}

NIMA and SAAN produce the most strongly correlated pair of maps anywhere in this study, agreeing across architectural class profiles at $r\approx0.98$. Taken at face value this would be the headline result of the paper, since two models trained on different data for different purposes appear to have converged on a common reading of architecture. The pipeline gives good reason to resist that reading.

Both models return one number per image and therefore pass through the segmentation stage of Section~\ref{sec:mask}, and they pass through the same one: identical rendered views are divided into identical regions, and each model writes a constant value across every region. Their outputs are piecewise-constant fields over the same partition of every view, so any structure the partition imposes is shared by construction. Because these regions tend to follow material and illumination boundaries, and because both scorers respond strongly to exposure, contrast, and texture density within a crop, much of the observed agreement is agreement about where the boundaries fall and how bright the enclosed surface is, which is a property of the segmenter and the renderer rather than of aesthetics. Table~\ref{tab:mask-ablation} supports this from another angle: on masked photographic crops, where both models can be compared against real human ratings, SAAN's correlation with AVA is indistinguishable from zero under every operator. Two models that agree closely with each other while only one agrees with human raters are not agreeing about human aesthetic judgment.

The same caution applies in principle to PE and SigLIP2, which also depend on the mask stage and turn out to be the most strongly agreeing pair among the text-query backbones. Section~\ref{sec:res-residual} shows why that case is different in practice.

\subsection{Text querying produces the most legible maps}
\label{sec:res-text}

\begin{figure}[tbp]
    \centering
  \includegraphics[width=0.85\textwidth]{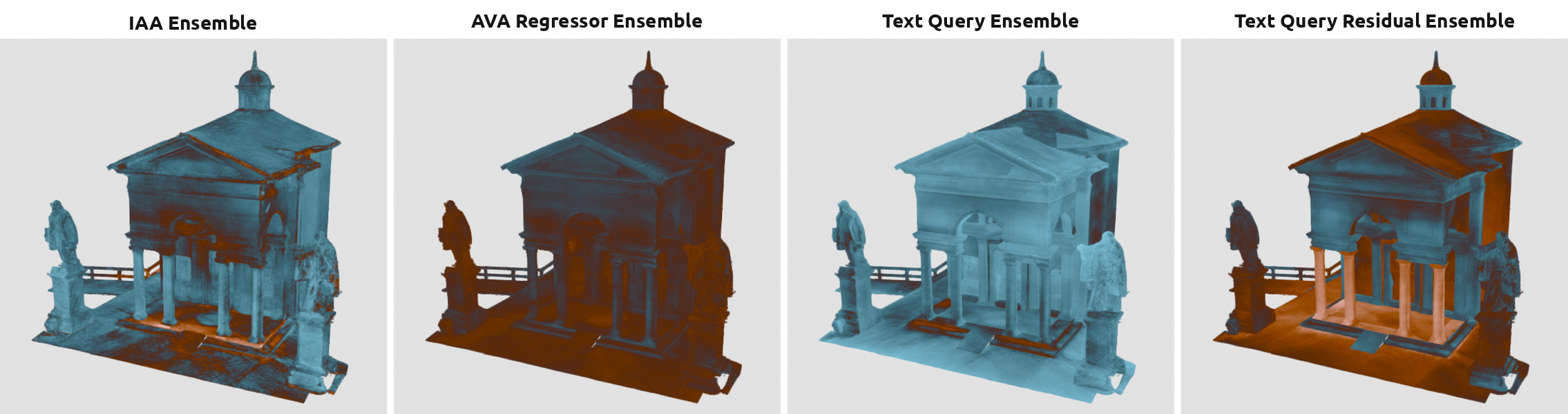}
  \caption{Route ensembles on ArCH \texttt{5\_SMV\_chapel\_1}. Left to right: mean of lifted SAAN and NIMA scores; mean of the five AVA regression heads; mean of the four raw text-query maps (DINOv3, TIPS, PE, SigLIP2); mean of the same four maps after class-text residualisation. Blue is high, red is low. The residual text-query ensemble produces sharper distinctions between objects, including between portico columns and inner facade columns that share a class label.}
  \label{fig:ens-arch}
\end{figure}

\begin{figure}[tbp]
    \centering
  \includegraphics[width=0.85\textwidth]{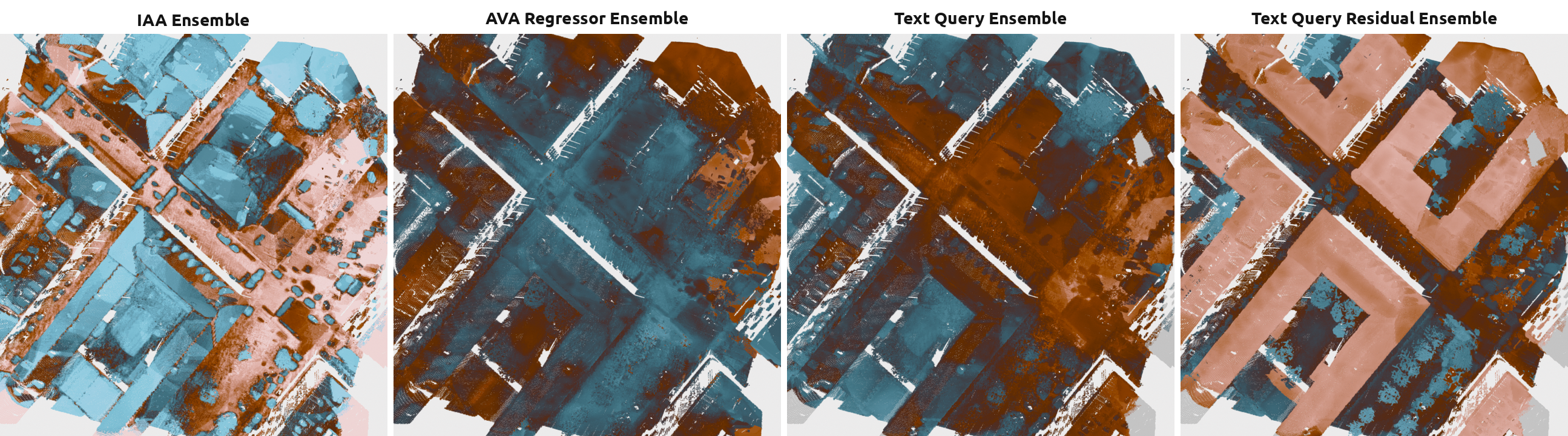}
  \caption{The same four route ensembles on Turin3D \texttt{scene\_14}. Within-route coherence does not imply agreement across routes. The residual text-query ensemble clearly prefers trees and grass over built objects such as cars and buildings, a distinction that is less apparent in the other routes.}
  \label{fig:ens-turin}
\end{figure}

Figures~\ref{fig:ens-arch} and~\ref{fig:ens-turin} summarise the three scoring routes as ensembles on a representative ArCH church and a Turin3D block. Across the text-query panels in particular, boundaries between high- and low-scoring regions tend to coincide with architectural boundaries rather than falling arbitrarily across surfaces, which indicates that the lifted features carry structural and material variation into the cloud. What differs between models, and between routes, is valence rather than structure: they largely agree about where the divisions in a building are and disagree about which side of a division to favour.

\begin{figure}[tbp]
    \centering
  \includegraphics[width=\textwidth]{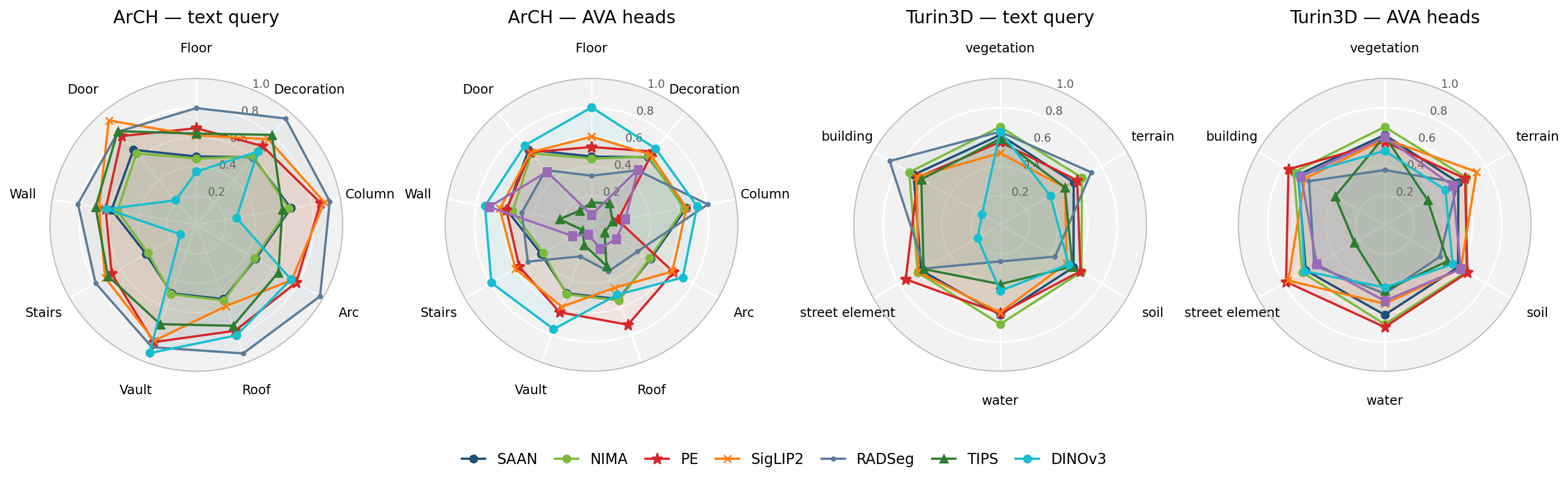}
  \caption{Aesthetic ratios by semantic class on ArCH and Turin3D. A high value means that a large fraction of the class's points score above $0.5$. Text-query panels use the ``beautiful object'' prompt; AVA panels use the regression heads. A uniformly high or low profile indicates indiscriminate scoring rather than superior performance.}
  \label{fig:radar}
\end{figure}

Figure~\ref{fig:radar} summarises the class profiles. DINOv3 produces the most differentiated reading, placing decoration above floor on ArCH and vault well above stairs, an ordering that is consistent with an architectural interpretation but does not establish one. RADSeg, the backbone that fails the semantic check, is near-uniformly high on ArCH. PE's and SigLIP's class ratios sit in a mid range and do discriminate, if modestly. The backbone with the strongest dense features and the best semantic transfer is also the one whose aesthetic profile is most legible in architectural terms, and the backbone that fails the semantic check is the one whose profile is least interpretable.

\subsection{Removing class semantics reveals shared aesthetic structure}
\label{sec:res-residual}

\begin{figure}[tbp]
    \centering
  \includegraphics[width=0.75\textwidth]{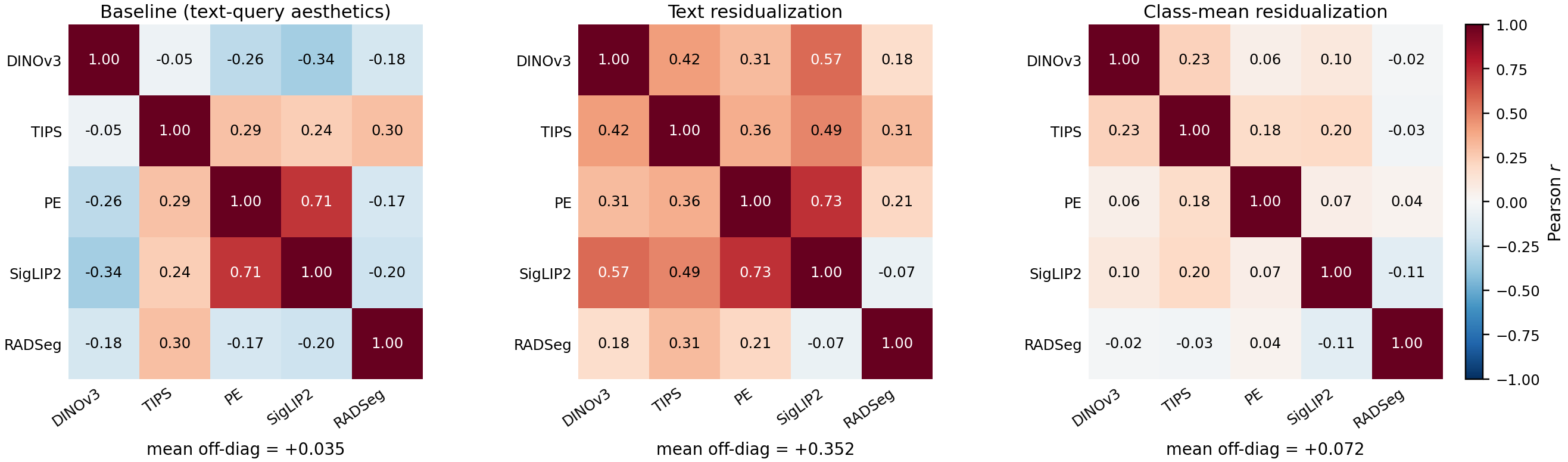}
  \caption{Cross-model correlation of per-class text-query aesthetic ratios on ArCH and Turin3D for DINOv3, TIPS, PE, SigLIP2, and RADSeg, Fisher-averaged over 53 scenes. Left, baseline scores. Centre, after projecting the class-text direction out of each point feature. Right, after projecting out the empirical class-mean feature. Only the language-mediated residualisation produces a substantial gain.}
  \label{fig:corr-residual}
\end{figure}

\begin{figure}[tbp]
    \centering
  \includegraphics[width=0.75\textwidth]{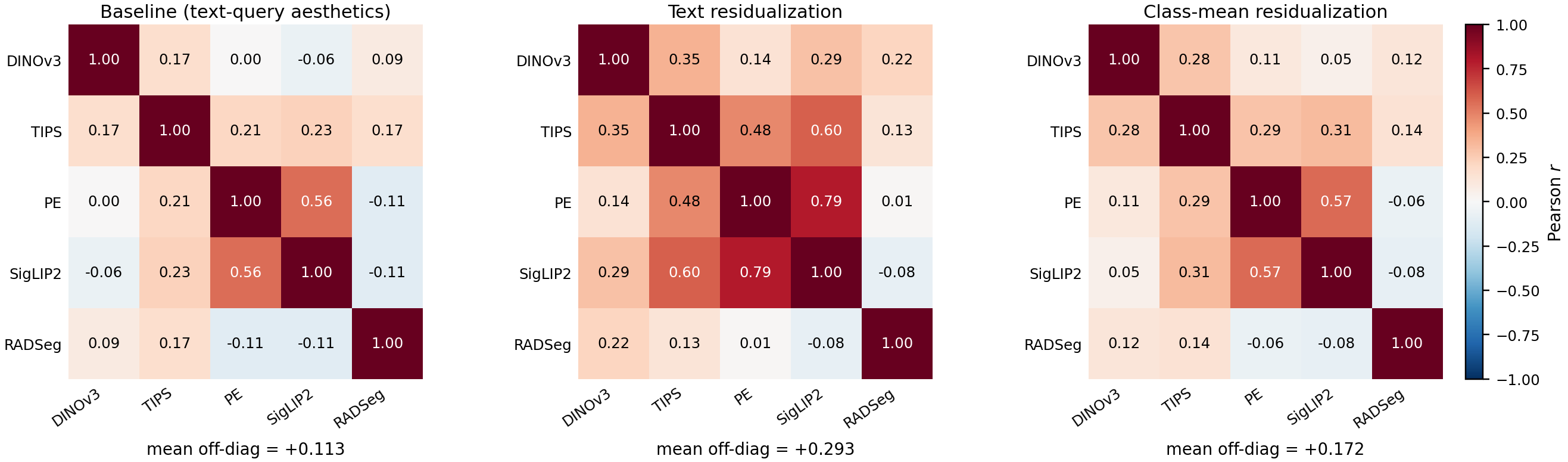}
  \caption{The same three protocols evaluated on per-point score vectors rather than class profiles. Baseline agreement is already positive for almost every pair, and residualisation raises the mean from $+0.113$ to $+0.293$.}
  \label{fig:corr-residual-pointwise}
\end{figure}

Restricted to the five text-query backbones, baseline agreement is slight over class profiles, with a mean off-diagonal correlation of $+0.035$ across ArCH and Turin3D, but clearer at the point level, at $+0.113$ (Figure~\ref{fig:corr-residual}, left; Figure~\ref{fig:corr-residual-pointwise}, left). Class profiles compare only a few metrics and are thus more fragile, whereas point-level comparison averages over millions of measurements.

A likely reason is that the baseline profiles measure semantics as well as aesthetics. If the direction meaning ``this is decoration'' sits closer to ``beautiful object'' than the direction meaning ``this is a floor'', the profile can report decoration as more beautiful for lexical rather than perceptual reasons. Because each backbone arranges its class directions differently, this contribution appears as disagreement.

We test this by removing the distracting vector direction. Let $x$ be an $\ell_2$-normalised lifted feature and $d_c$ the unit text embedding of the point's ground-truth class. We replace $x$ by its component orthogonal to $d_c$,
\begin{equation}
\tilde{x}
=
\frac{x-(x\cdot d_c)\,d_c}{\lVert x-(x\cdot d_c)\,d_c\rVert_2},
\end{equation}
and recompute the aesthetic score as cosine similarity to the same prompt, renormalised to $[0,1]$. This prevents the score from reporting how closely a point resembles its class name. As a control, we subtract each class's empirical mean feature instead of its text direction.

Mean off-diagonal agreement rises from $+0.035$ to $+0.352$ over class profiles and from $+0.113$ to $+0.293$ over points. The class-mean control reaches only $+0.072$ and $+0.172$, indicating that the gain is specific to removing the class-text direction rather than class-predictive variance in general.

The pattern of cross-model agreement is just as informative as the average, if not more so. Over class profiles, every pair among DINOv3, TIPS, PE, and SigLIP2 is positive, ranging from $0.31$ for DINOv3 with PE to $0.73$ for PE with SigLIP2, and three of the six were negative before residualisation. RADSeg is the exception, agreeing with the others between $-0.07$ and $0.31$, and it is also the backbone that failed the semantic check of Table~\ref{tab:miou} and the AVA regression of Table~\ref{tab:regressors}. Agreement tracks representational quality: the models that can distinguish urban and architectural semantics converge on how to score them, but the semantically incoherent RADSeg does not. The point-level matrix shows the same ordering: PE and SigLIP2 agree at $0.79$, while TIPS and SigLIP2 agree at $0.60$.

This also disposes of the concern raised at the end of Section~\ref{sec:res-iaa}. Were the agreement among text-query models an artefact of the shared mask stage, it would be confined to PE and SigLIP2, the two models that use it. Instead the two dense backbones, which never touch a segmenter, agree with one another at $0.42$ and with the masked pair at between $0.31$ and $0.57$. Agreement crosses the boundary between the two extraction regimes, which it could not do if segmentation alone produced it, although part of the high agreement between SigLIP2 and PE is certainly attributable to their shared masks.

The asymmetry between this section and the last is the practical finding of the paper. Text querying yields a signal that five independently trained backbones partially share once the semantic confound is removed; scalar IAA lifting yields a signal its two members share for reasons largely traceable to the segmenter; but AVA regression yields a signal its own members do not share. Most concerningly, no two routes reproduce each other.

\subsection{Disagreement between aesthetic labeling routes}
\label{sec:res-cross}

\begin{figure}[H]
  \centering
  \sbox{\crossrouteboxb}{\includegraphics[width=0.62\textwidth]{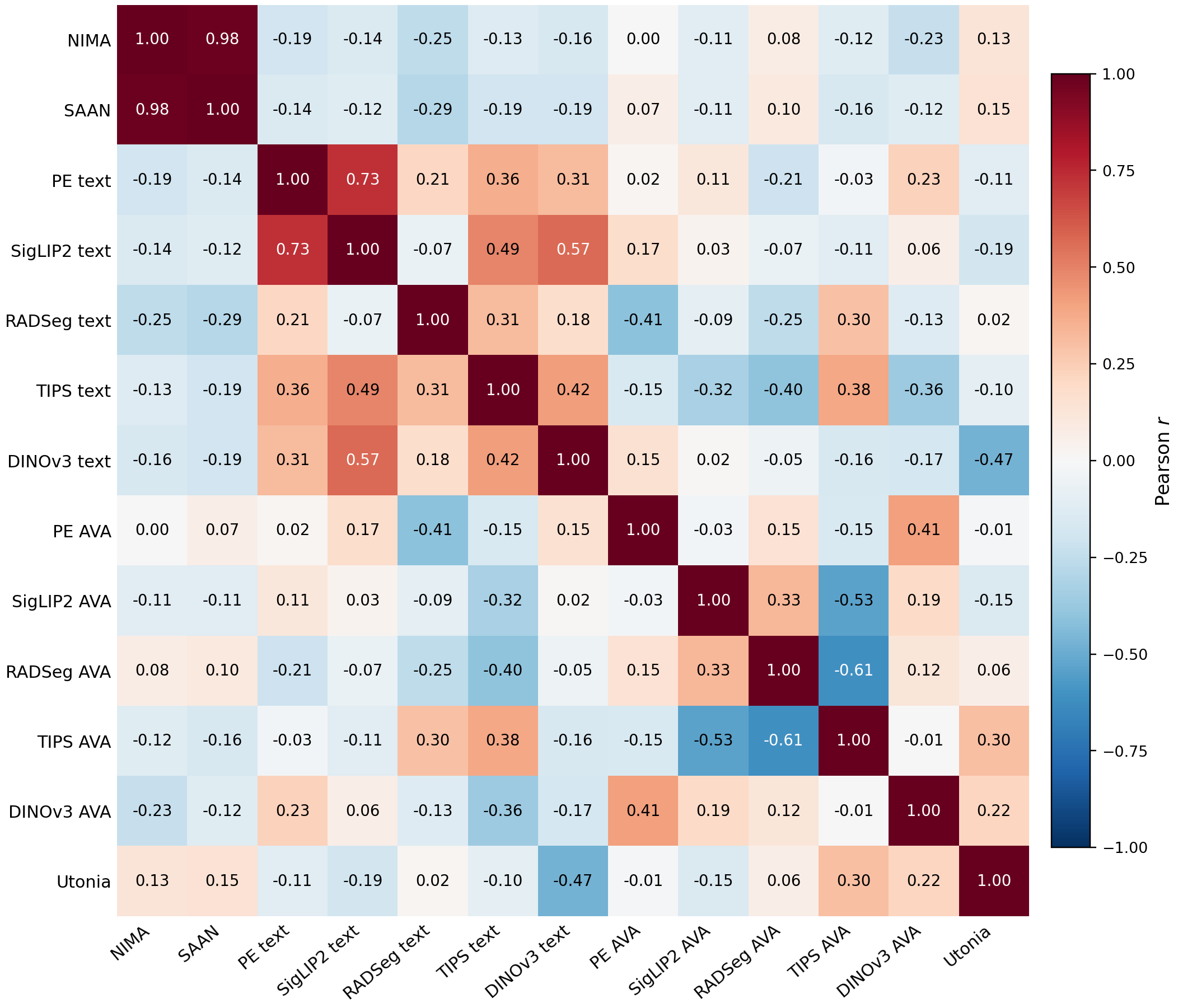}}
  \begin{minipage}[c][\ht\crossrouteboxb][c]{0.34\textwidth}
    \centering
    \includegraphics[width=\linewidth]{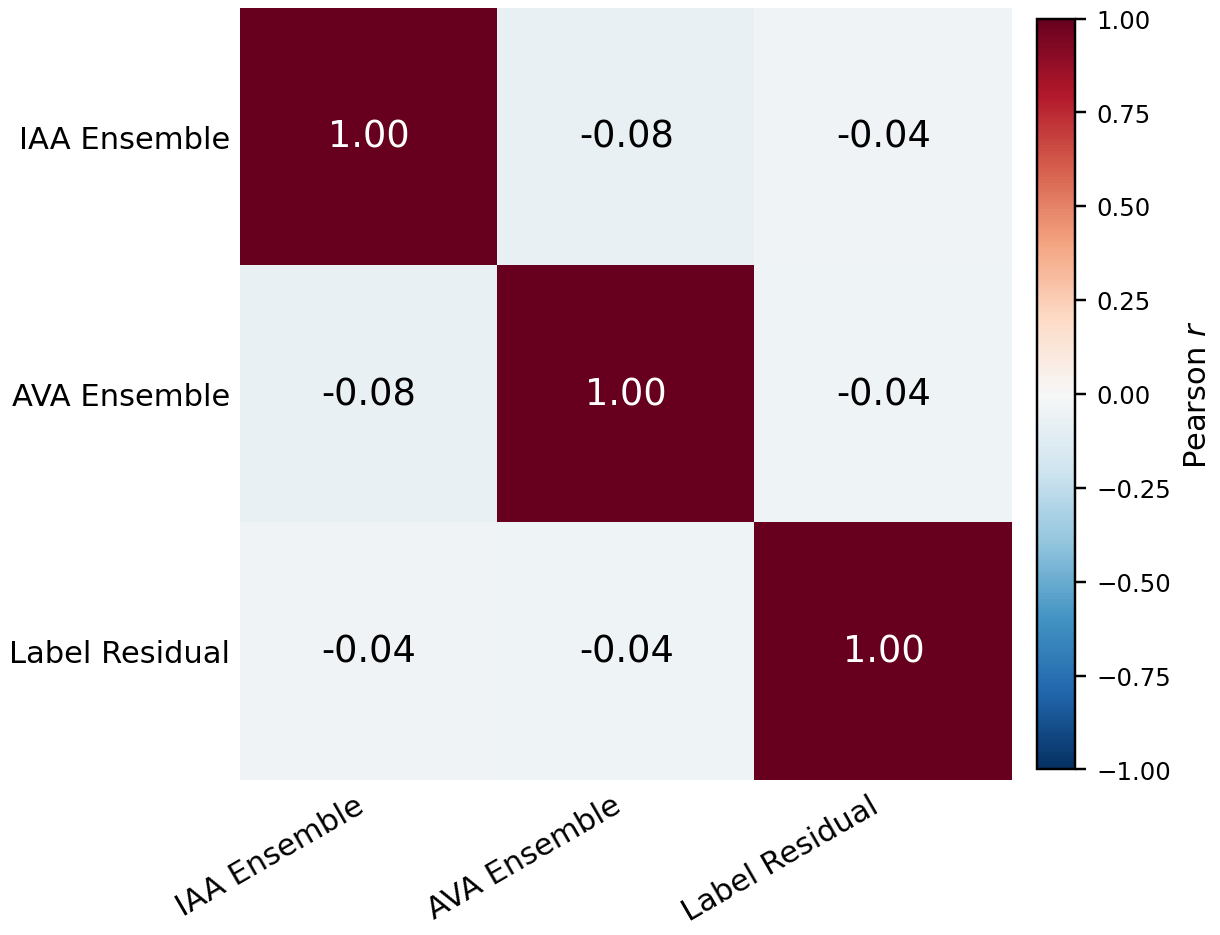}
  \end{minipage}\hfill
  \begin{minipage}[c][\ht\crossrouteboxb][c]{0.62\textwidth}
    \centering
    \usebox{\crossrouteboxb}
  \end{minipage}\\[0.5em]
  \captionsetup{type=figure}
  \begin{minipage}[t]{0.34\textwidth}
    \subcaption{Pointwise Pearson correlation between residual route ensembles (IAA, AVA, label-residual), Fisher-averaged. Near-zero off-diagonals ($-0.08$ to $-0.04$).}
    \label{fig:route-ensemble-pointwise-corr}
  \end{minipage}\hfill
  \begin{minipage}[t]{0.62\textwidth}
    \subcaption{Per-class aesthetic-ratio correlation across thirteen configurations on ArCH and Turin3D, Fisher-averaged over 53 scenes. Text-query columns are class-text residualised; others are lifted IAA, AVA, and Utonia.}
    \label{fig:corr}
  \end{minipage}
  \caption{Cross-route disagreement. Left: residual route ensembles at the point level. Right: full per-class correlation matrix.}
  \label{fig:cross-route}
\end{figure}

Convergence among the text-query backbones is convergence within one route, and it does not extend across the three. Averaging the five residualised text-query maps into an ensemble and correlating it against the other families gives values close to zero at the point level on both datasets, with Fisher-pooled correlations between $-0.04$ and $+0.05$. Class-profile correlations are similarly weak, at $-0.14$ against the ensemble of lifted NIMA and SAAN scores, $-0.08$ against the ensemble of AVA regression heads, and $+0.01$ against the Utonia probe. The one moderate exception is instructive rather than encouraging: the AVA adapters agree with the text-query ensemble at $+0.35$ on ArCH and disagree at $-0.23$ on Turin3D, a reversal between a heritage dataset and an urban one that is hard to read as anything but instability. Figure~\ref{fig:cross-route} summarises this: pointwise residual ensembles (a) and the full per-class matrix across all thirteen scoring configurations (b).

\FloatBarrier
\section{Discussion}
\label{sec:discussion}

\paragraph{Agreement is not validation.}
Five backbones converging on a reading of a building shows stability across architectures, objectives, training data, and representational geometry. It is not evidence of accuracy: no people have yet judged these clouds for aesthetic content, and projecting out the class-text direction says nothing about agreement with human preference.

\paragraph{Proposed human 3D aesthetic assessment protocol.}
The missing evaluation is human-derived, part-aware judgment on these clouds. A sufficient protocol would use scenes at building and city scale, provide renders of regions carrying multiple class labels, and ask heritage-aware raters which region is more aesthetically salient. Continuous pointwise scores could cover color, material, harmony with other building elements, and general aesthetic quality. Rater judgments would be compared with each route's scores and could supervise a 3D-native regressor, providing a benchmark for future automatic 3D aesthetic assessment.

\paragraph{The domain gap runs deeper than rendering.}
Rendered views substitute for photographs well enough, as Section~\ref{sec:render} shows. The larger gap is between photographic and architectural judgment. AVA asks whether an image is a good photograph; a conservator asks whether a surface is significant, whether an intervention is legible, or whether a repair respects the fabric. Riegl and Brandi describe values in which a weathered, unphotogenic wall may be the most important element in a scene \cite{riegl1982monuments,brandi2005theory}. Photographic models tend to reward lighting, texture, and framing; color on ArCH and Turin3D mitigates but does not remove this problem.

\paragraph{Why disagreement between routes is not by itself a failure.}
Averaging the three routes would produce a smoother field, but without an external target it would be no closer to human judgment and would conceal which aesthetic culture generated it. The proposed human protocol should therefore evaluate each route separately.

\section{Conclusion}
\label{sec:conclusion}

We compared three ways of carrying image aesthetic assessment onto scanned buildings and cities. Learning the mapping from human ratings proved least durable: AVA regression heads reproduce photographic judgment within photography but lose coherence on architectural renders, while the 3D-native probe stays close to chance because of reconstruction quality. Lifted IAA scalars agree closely, but largely because the two models share a segmentation stage.

Open-vocabulary text querying carries the most coherent signal. Agreement is positive before correction and rises after projecting out the class-text direction, reaching $+0.352$ over architectural part profiles and $+0.293$ over individual points. It holds among the four backbones that transfer semantic structure into 3D and crosses dense and mask-based extraction, ruling out the segmenter as its source. Residualisation separates aesthetic response from semantic identity and should be standard in future evaluative queries.

Whether these maps correspond to human aesthetic judgments in 3D remains unknown. Until part-aware human validation is carried out, they should be presented as reproducible computational hypotheses rather than measurements.

\section*{AI Use Reflection}
Literature organisation and wording assistance used a coding agent under author supervision. Empirical numbers and scholarly claims were checked against code, logs, and sources by the authors, who remain responsible for their correctness.

\printbibliography

\appendix

\section{Implementation details}
\label{app:train}

\textbf{Segmentation.} We use the SAM 2 automatic mask generator with a ViT-H checkpoint \cite{ravi2024sam2}. Overlapping masks are not resolved by exclusive assignment; embeddings and scores are unrolled with mask-confidence weighting, so contested pixels receive blended contributions from every region that claims them. We do not sweep the density and stability thresholds, which govern region granularity and would therefore alter the maps produced by the global models. Ablation renders for the context operators were generated under the same segmenter using a denser selection of the largest regions per image.

\textbf{2D AVA heads.} A two-layer perceptron with linear, GELU, layer-normalisation, and linear stages, trained with Adam at $1\times10^{-3}$ for 10 epochs on the full AVA dataset. Each patch or mask embedding inherits the global score of its source image.

\textbf{Utonia probe.} Up to 50 epochs with AdamW at $1\times10^{-4}$ and early stopping after five validation steps without improvement. The loss combines point-wise binary cross-entropy on scores binarised at $0.5$ with a batch-level histogram Kullback--Leibler term, $\mathcal{L}=\mathcal{L}_{\mathrm{BCE}}+\lambda\mathcal{L}_{\mathrm{KL}}$ with $\lambda=0.5$ in the reported run, where each scene is summarised by its mean predicted probability and mean ground-truth score. Points inherit scene-level AVA scores from the MapAnything reconstructions.

\textbf{Prompt ablation.} We compared six positive prompts under cosine scoring and also evaluated negative and neutral contrasts for the primary prompt. The table reports mean off-diagonal agreement over the two representative scenes used in the ablation; ``text residual'' projects out each point's class-text direction before rescoring. The contrast formulations were substantially more variable than direct cosine similarity, supporting the fixed ``beautiful object'' choice for the main experiments.

\begin{table}[H]
\centering
\caption{Prompt ablation on ArCH \texttt{5\_SMV\_chapel\_1} and Turin3D \texttt{scene\_14}. Values are Fisher-pooled mean off-diagonal Pearson correlations across the five text-query backbones, reported for class profiles and pointwise scores. The positive-prompt rows use cosine similarity; the final rows use ``beautiful object'' with a negative or neutral contrast. Direct cosine scoring is more stable than the contrast formulations across the two scenes and is used thereafter.}
\label{tab:prompt-ablation}
\scriptsize
\setlength{\tabcolsep}{3pt}
\begin{tabular}{lrrrr}
\toprule
Prompt / protocol & Class & Class residual & Point & Point residual \\
\midrule
\texttt{beautiful object} & $+0.181$ & $+0.427$ & $+0.111$ & $+0.336$ \\
\texttt{aesthetically pleasing object} & $+0.155$ & $+0.409$ & $+0.086$ & $+0.336$ \\
\texttt{beautiful scene} & $-0.003$ & $+0.444$ & $+0.094$ & $+0.284$ \\
\texttt{aesthetically pleasing scene} & $+0.005$ & $+0.424$ & $+0.047$ & $+0.290$ \\
\texttt{beautiful view} & $+0.132$ & $+0.409$ & $+0.126$ & $+0.240$ \\
\texttt{aesthetically pleasing view} & $-0.106$ & $+0.372$ & $+0.056$ & $+0.299$ \\
\midrule
\texttt{beautiful object}, negative contrast & $-0.382$ & $-0.343$ & $-0.003$ & $+0.010$ \\
\texttt{beautiful object}, neutral contrast & $-0.256$ & $-0.234$ & $+0.060$ & $+0.022$ \\
\bottomrule
\end{tabular}
\end{table}

\section{MapAnything reconstruction and 3D probing}
\label{app:mapanything}

\begin{figure}[H]
\centering
  \begin{subfigure}{0.28\linewidth}
    \includegraphics[width=\linewidth]{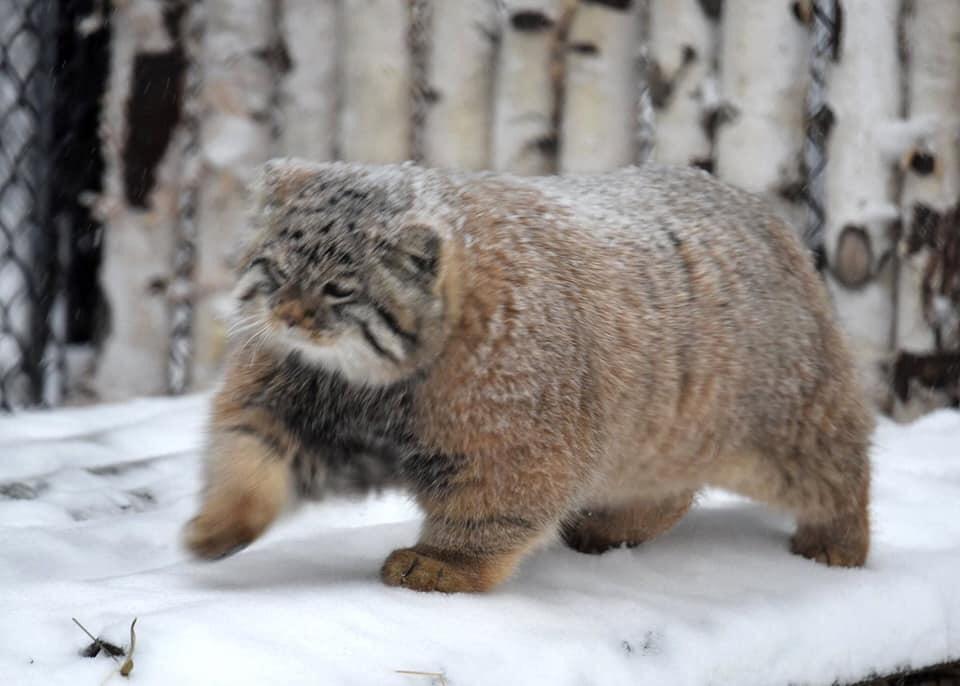}
    \caption{Original image}
  \end{subfigure}\hfill
  \begin{subfigure}{0.28\linewidth}
    \includegraphics[width=\linewidth]{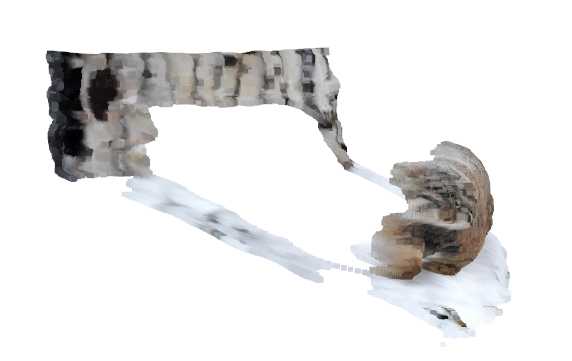}
    \caption{MapAnything reconstruction}
  \end{subfigure}\hfill
  \begin{subfigure}{0.28\linewidth}
    \includegraphics[width=\linewidth]{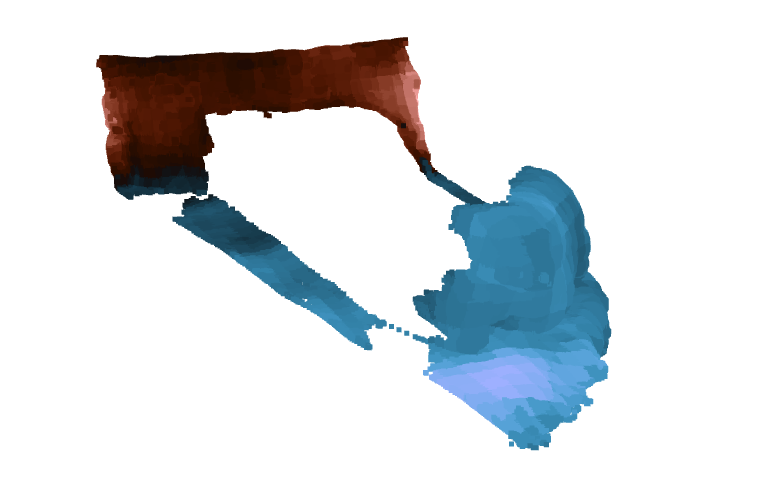}
    \caption{Utonia aesthetic probe}
  \end{subfigure}
  \caption{Toy visualisation of the MapAnything--Utonia route. Coarse scene identity is preserved under depth distortion; the probe still yields locally distinct predictions under global supervision (representational localisation, not metric fidelity).}
  \label{fig:mapanything-cat}
\end{figure}

Figure~\ref{fig:mapanything-cat} makes the geometric limit of the 3D-native route concrete. MapAnything recovers the broad subject--background layout, but monocular depth is stretched and folded: the animal is displaced and the supporting surface becomes elongated and fragmented. Supervision remains global---every reconstructed point inherits its source photograph's AVA score---yet frozen Utonia features and the attentive probe still yield spatially distinct predictions. Those local differences show what the probe can read from the representation, not metric fidelity or human-validated aesthetics.

\section{Additional visualisations}
\label{app:viz}

\begin{figure}[p]
\centering
\begin{adjustbox}{max width=\textwidth, max totalheight=0.84\textheight, center}
\begin{minipage}{\textwidth}
\centering
  \begin{subfigure}{0.32\linewidth}
    \includegraphics[width=\linewidth]{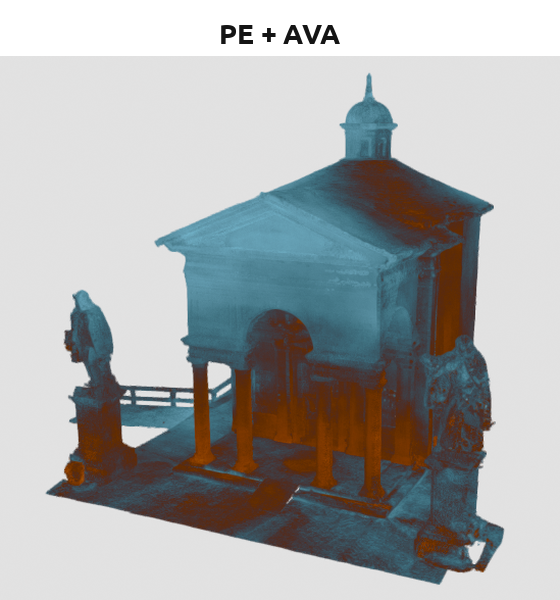}
  \end{subfigure}\hfill
  \begin{subfigure}{0.32\linewidth}
    \includegraphics[width=\linewidth]{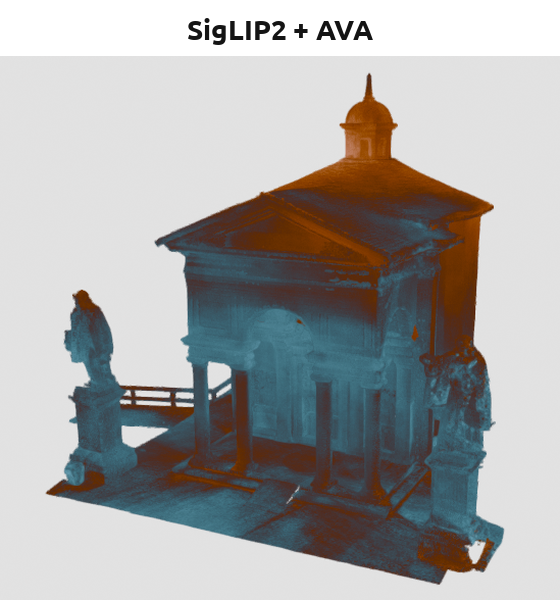}
  \end{subfigure}\hfill
  \begin{subfigure}{0.32\linewidth}
    \includegraphics[width=\linewidth]{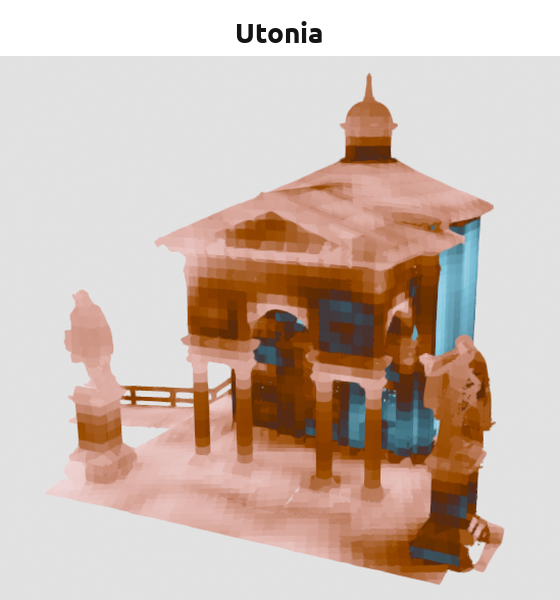}
  \end{subfigure}
  \par\vspace{2pt}\noindent
  \begin{subfigure}{0.32\linewidth}
    \includegraphics[width=\linewidth]{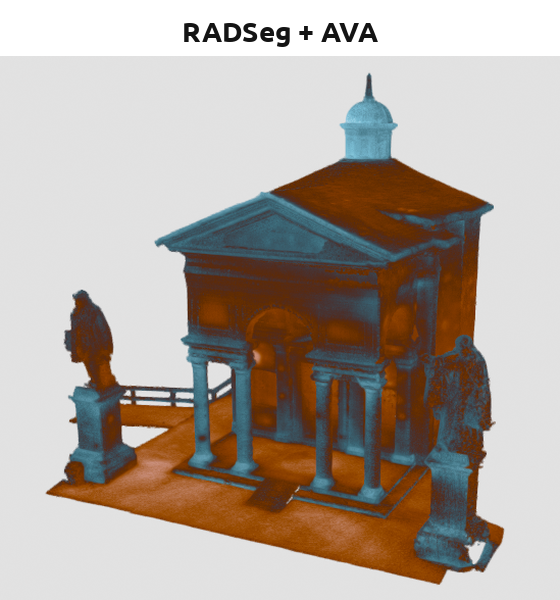}
  \end{subfigure}\hfill
  \begin{subfigure}{0.32\linewidth}
    \includegraphics[width=\linewidth]{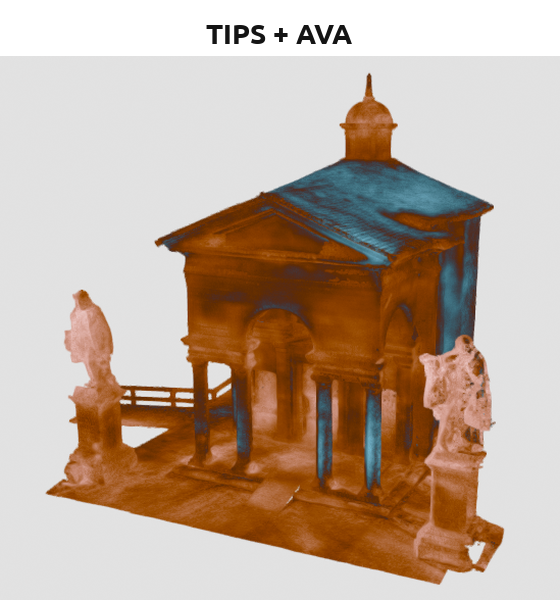}
  \end{subfigure}\hfill
  \begin{subfigure}{0.32\linewidth}
    \includegraphics[width=\linewidth]{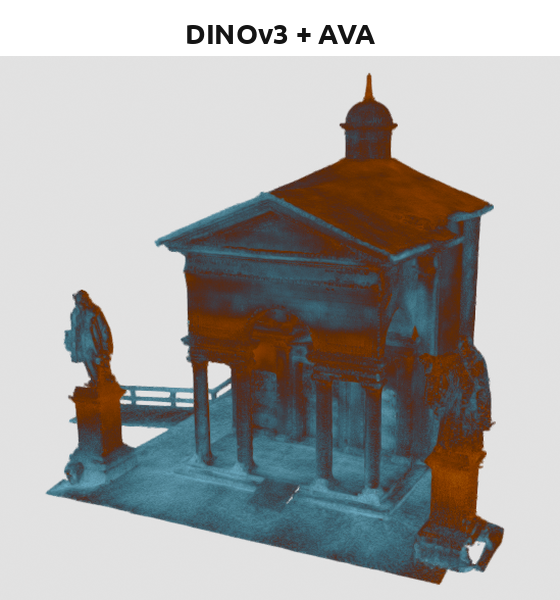}
  \end{subfigure}
  \par\vspace{2pt}\noindent
  \begin{subfigure}{0.32\linewidth}
    \includegraphics[width=\linewidth]{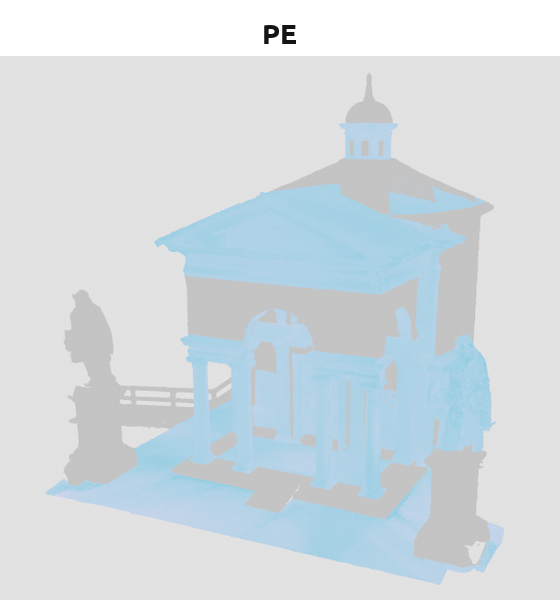}
  \end{subfigure}\hfill
  \begin{subfigure}{0.32\linewidth}
    \includegraphics[width=\linewidth]{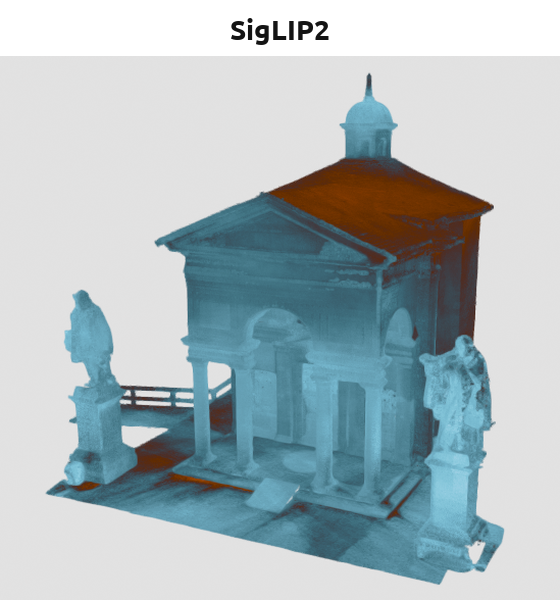}
  \end{subfigure}\hfill
  \begin{subfigure}{0.32\linewidth}
    \includegraphics[width=\linewidth]{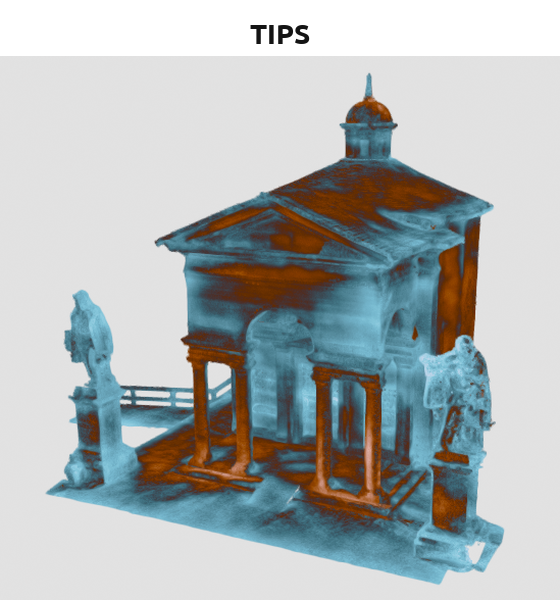}
  \end{subfigure}
  \par\vspace{2pt}\noindent
  \begin{subfigure}{0.32\linewidth}
    \includegraphics[width=\linewidth]{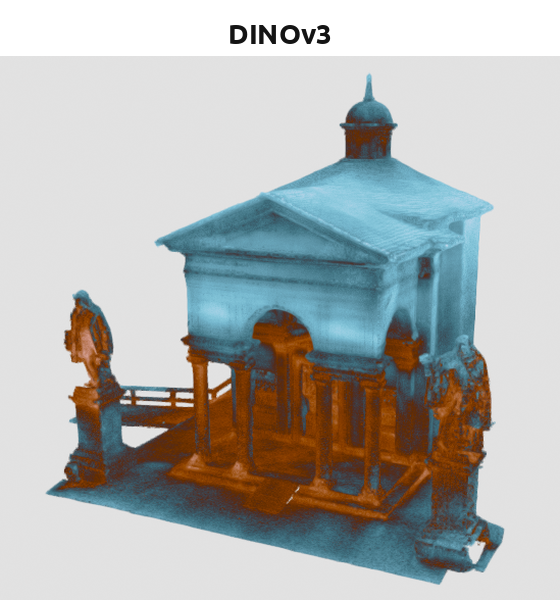}
  \end{subfigure}\hfill
  \begin{subfigure}{0.32\linewidth}
    \includegraphics[width=\linewidth]{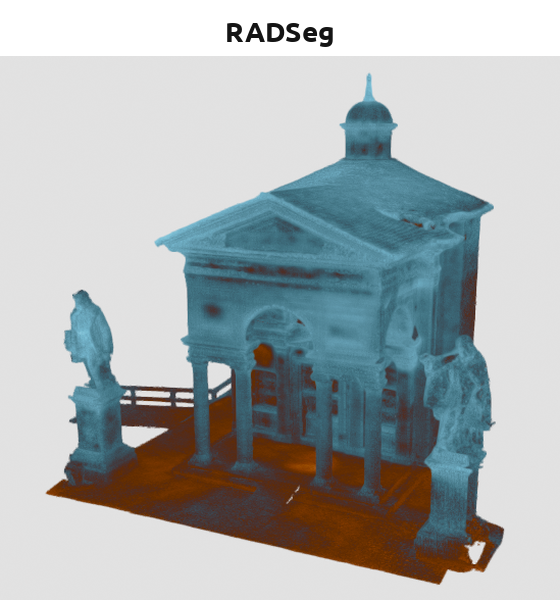}
  \end{subfigure}\hfill
  \hspace{0.32\linewidth}
\end{minipage}
\end{adjustbox}
  \caption{ArCH aesthetic maps for the AVA regression heads, Utonia probe, and raw text-query models on the same church and camera as Figure~\ref{fig:ens-arch}. The native labels above each render identify every panel; the rows continue sequentially without empty slots.}
  \label{fig:ava-and-label-maps}
\end{figure}

\begin{figure}[p]
\centering
\begin{adjustbox}{max width=\textwidth, max totalheight=0.84\textheight, center}
\begin{minipage}{\textwidth}
\centering
  \begin{subfigure}{0.24\linewidth}
    \includegraphics[width=\linewidth]{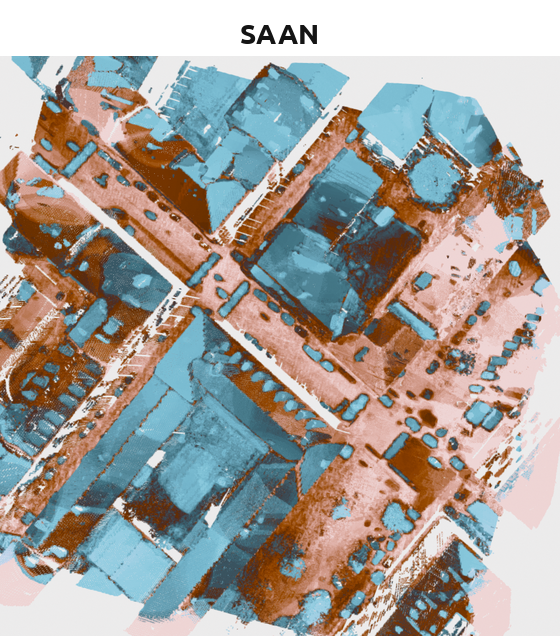}
  \end{subfigure}\hfill
  \begin{subfigure}{0.24\linewidth}
    \includegraphics[width=\linewidth]{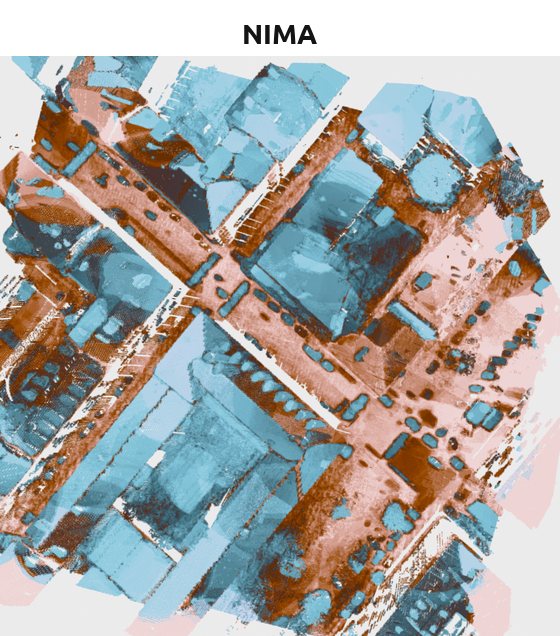}
  \end{subfigure}\hfill
  \begin{subfigure}{0.24\linewidth}
    \includegraphics[width=\linewidth]{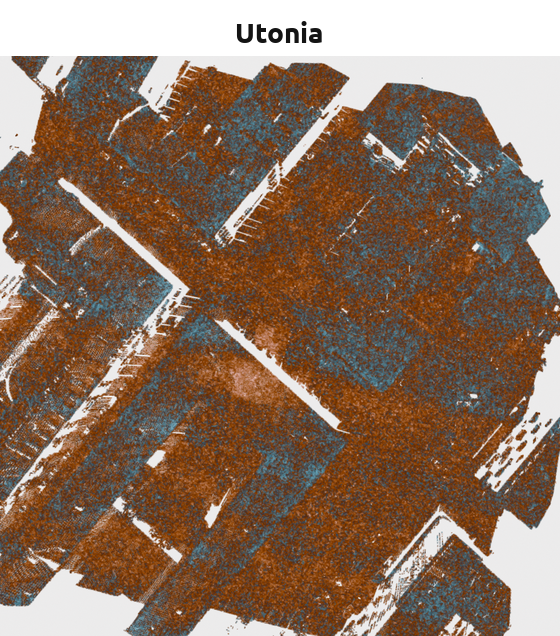}
  \end{subfigure}\hfill
  \begin{subfigure}{0.24\linewidth}
    \includegraphics[width=\linewidth]{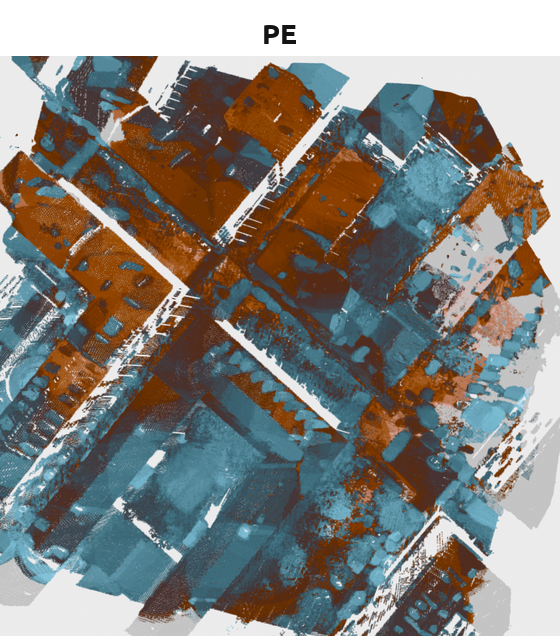}
  \end{subfigure}
  \par\vspace{2pt}\noindent
  \begin{subfigure}{0.24\linewidth}
    \includegraphics[width=\linewidth]{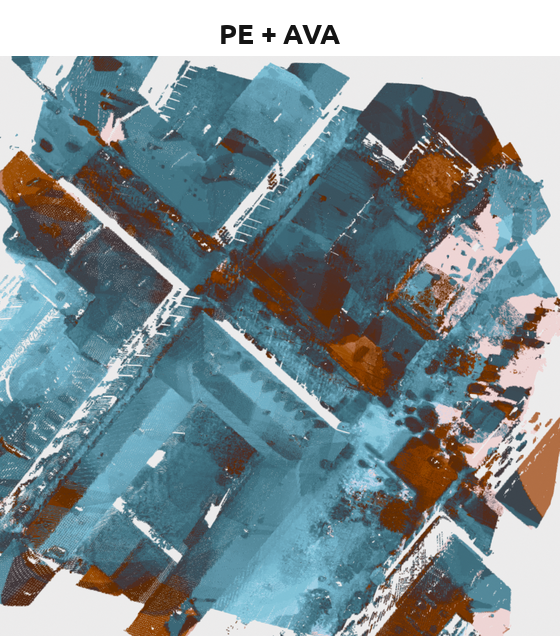}
  \end{subfigure}\hfill
  \begin{subfigure}{0.24\linewidth}
    \includegraphics[width=\linewidth]{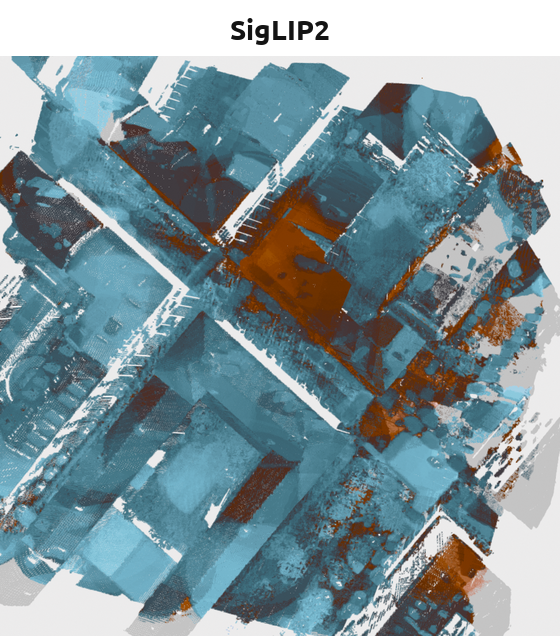}
  \end{subfigure}\hfill
  \begin{subfigure}{0.24\linewidth}
    \includegraphics[width=\linewidth]{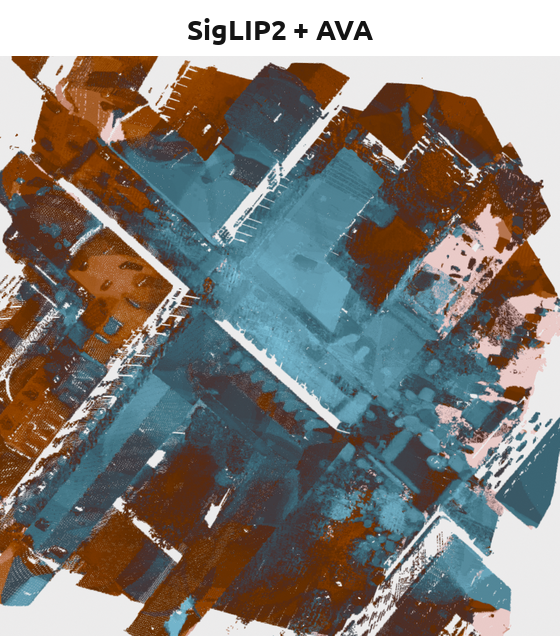}
  \end{subfigure}\hfill
  \begin{subfigure}{0.24\linewidth}
    \includegraphics[width=\linewidth]{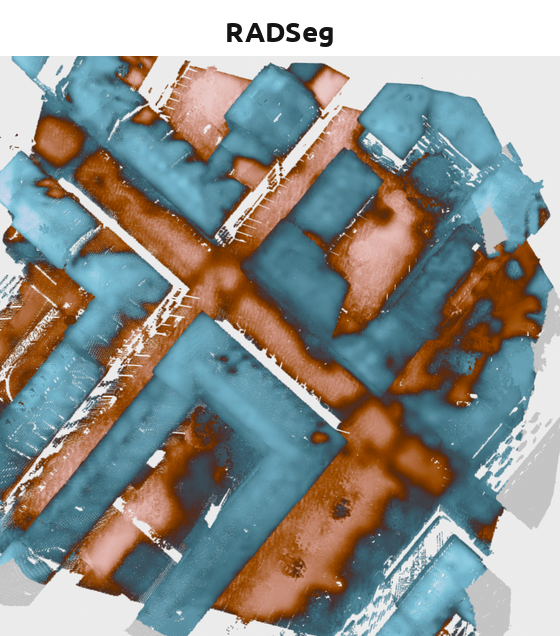}
  \end{subfigure}
  \par\vspace{2pt}\noindent
  \begin{subfigure}{0.24\linewidth}
    \includegraphics[width=\linewidth]{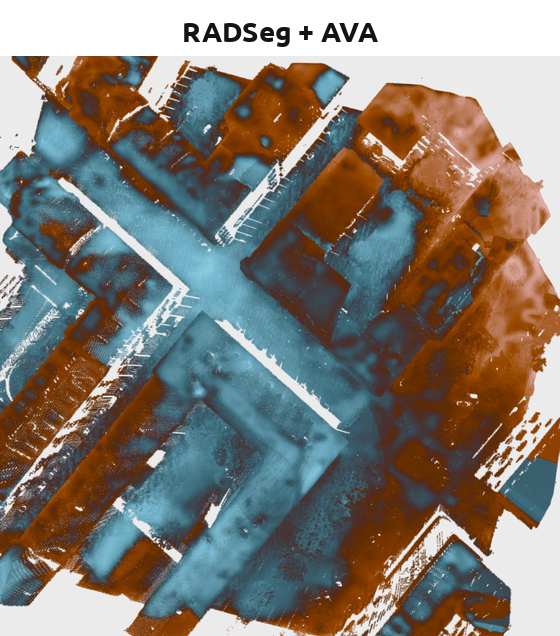}
  \end{subfigure}\hfill
  \begin{subfigure}{0.24\linewidth}
    \includegraphics[width=\linewidth]{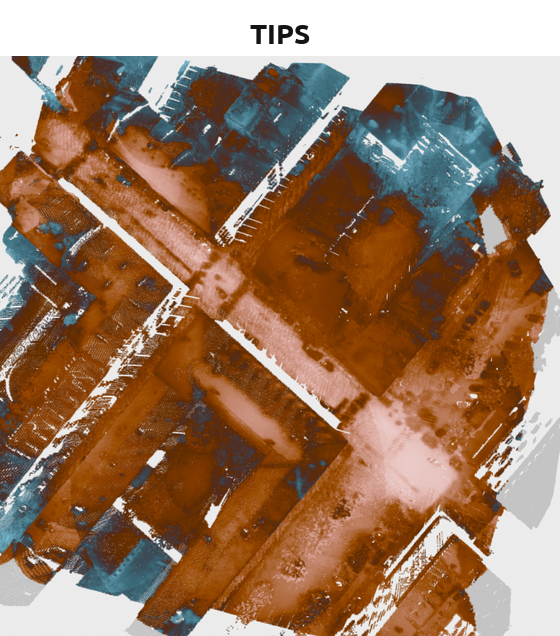}
  \end{subfigure}\hfill
  \begin{subfigure}{0.24\linewidth}
    \includegraphics[width=\linewidth]{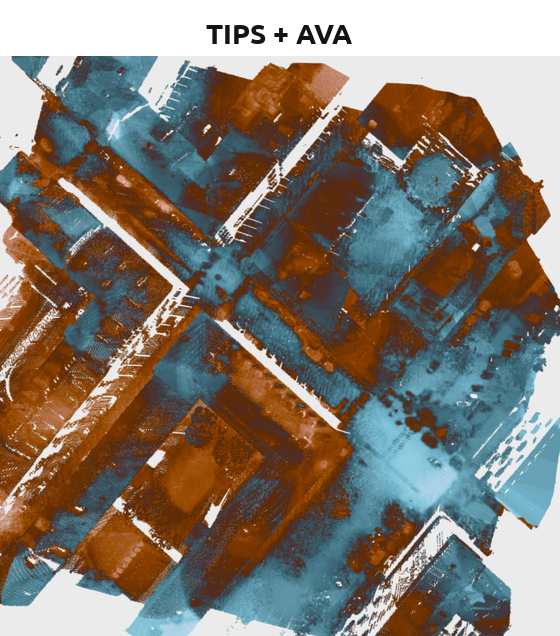}
  \end{subfigure}\hfill
  \begin{subfigure}{0.24\linewidth}
    \includegraphics[width=\linewidth]{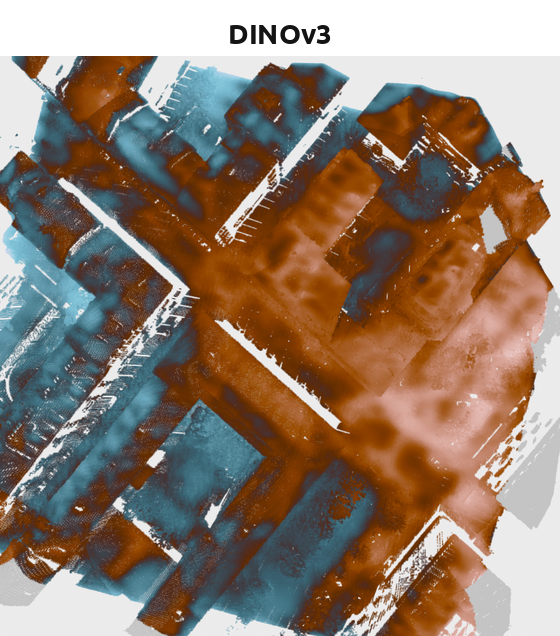}
  \end{subfigure}
  \par\vspace{2pt}\noindent
  \begin{subfigure}{0.24\linewidth}
    \includegraphics[width=\linewidth]{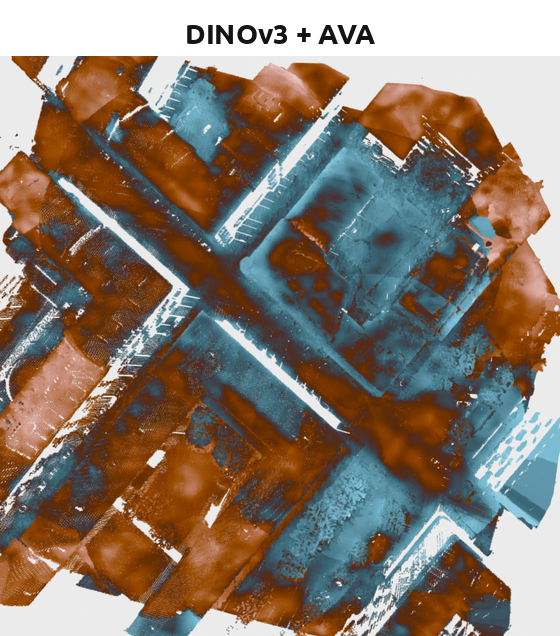}
  \end{subfigure}
\end{minipage}
\end{adjustbox}
  \caption{Aesthetic maps on Turin3D \texttt{scene\_14} (same camera and framing as Figure~\ref{fig:ens-turin}): lifted IAA scalars, the Utonia probe, and text-query / AVA maps. Each render carries its model label natively, and the rows continue sequentially without empty slots.}
  \label{fig:turin-top-maps}
\end{figure}

\end{document}